\documentclass{article}

\usepackage{microtype}
\usepackage{graphicx}
\usepackage{subfigure}
\usepackage{booktabs} 

\usepackage{hyperref}

\usepackage[accepted]{icml2025}

\usepackage{amsmath}
\usepackage{amssymb}
\usepackage{mathtools}
\usepackage{amsthm}

\usepackage[capitalize,noabbrev]{cleveref}

\theoremstyle{plain}

\theoremstyle{definition}

\theoremstyle{remark}

\usepackage[textsize=tiny]{todonotes}

\usepackage{ulem}
\usepackage{xspace}
\newcommand{\dataset}{\textsc{MME-CoT}\xspace}
\usepackage{xcolor}

\definecolor{mycolor_green}{HTML}{E6F8E0}
\definecolor{backred}{RGB}{255, 190, 190}
\definecolor{red}{RGB}{139, 0, 0}
\definecolor{purple}{HTML}{E6F8E0}

\definecolor{verylightgray}{HTML}{E6F8E0} 
\definecolor{lightgray}{gray}{0.95} 

\usepackage{multirow}
\usepackage{colortbl}
\usepackage{makecell}
\usepackage{caption}
\usepackage{adjustbox}
\makeatletter
  \newcommand\figcaption{\def\@captype{figure}\caption}
  \newcommand\tabcaption{\def\@captype{table}\caption}
\makeatother
\usepackage[most]{tcolorbox} 
\DeclareMathOperator*{\argmax}{arg\,max}
\usepackage{tabularx}

\icmltitlerunning{MME-CoT: Benchmarking Chain-of-Thought in LMMs for Reasoning Quality, Robustness, and Efficiency}

\begin{document}

\twocolumn[
\icmltitle{MME-CoT: Benchmarking Chain-of-Thought in Large Multimodal Models\\for Reasoning Quality, Robustness, and Efficiency}






\icmlsetsymbol{equal}{*}
\begin{icmlauthorlist}
\icmlauthor{Dongzhi Jiang$^{*}$}{CUHK MMLab}\textbf{, }
\icmlauthor{Renrui Zhang$^{*\dagger}$}{CUHK MMLab}\textbf{, }
\icmlauthor{Ziyu Guo}{CUHK MiuLar Lab}\textbf{, }
\icmlauthor{Yanwei Li$^{\ddagger}$}{ByteDance}\textbf{, }
\icmlauthor{Yu Qi$^{\ddagger}$}{Northeastern University}\textbf{, }
\icmlauthor{Xinyan Chen$^{\ddagger}$}{CUHK MMLab}\\
\icmlauthor{Liuhui Wang$^{\ddagger}$}{UPenn}\textbf{, }
\icmlauthor{Jianhan Jin$^{\ddagger}$}{NJU}\textbf{, }
\icmlauthor{Claire Guo$^{\ddagger}$}{CUHK (Shenzhen)}\textbf{, }
\icmlauthor{Shen Yan}{ByteDance}\textbf{, }
\icmlauthor{Bo Zhang}{Shanghai AI Laboratory}\\
\icmlauthor{Chaoyou Fu}{NJU}\textbf{, }
\icmlauthor{Peng Gao}{Shanghai AI Laboratory}\textbf{, }
\icmlauthor{Hongsheng Li}{CUHK MMLab}\\
\vspace{0.2cm}
\begin{tabular}{c}
\textsuperscript{1} CUHK MMLab\quad
\textsuperscript{2} CUHK MiuLar Lab\quad
\textsuperscript{3} ByteDance\quad
\textsuperscript{4} NEU\quad
\textsuperscript{5} UPenn\\
\textsuperscript{6} NJU\quad
\textsuperscript{7} CUHK (Shenzhen)\quad
\textsuperscript{8} Shanghai AI Laboratory
\\
\vspace{0.5em}
\texttt{\{dzjiang,renruizhang\}@link.cuhk.edu.hk}\\
\vspace{0.1cm}
$^{*}$ Core contribution\quad$^{\dagger}$ Project lead\quad$^{\ddagger}$ Equal contribution
   \end{tabular} 
\\
Project Page: \url{https://mmecot.github.io/}

\end{icmlauthorlist}




\icmlcorrespondingauthor{Dongzhi Jiang, Renrui Zhang}{dzjiang, renruizhang@link.cuhk.edu.hk}

\vskip 0.3in
]

\begin{abstract}
Answering questions with Chain-of-Thought (CoT) has significantly enhanced the reasoning capabilities of Large Language Models (LLMs), yet its impact on Large Multimodal Models (LMMs) still lacks a systematic assessment and in-depth investigation. In this paper, we introduce \textbf{MME-CoT}, a specialized benchmark evaluating the CoT reasoning performance of LMMs, spanning six domains: math, science, OCR, logic, space-time, and general scenes. 
As the first comprehensive study in this area, we propose a thorough evaluation suite incorporating three novel metrics that assess the reasoning quality, robustness, and efficiency at a fine-grained level.
Leveraging curated high-quality data and a unique evaluation strategy, we conduct an in-depth analysis of state-of-the-art LMMs, uncovering several key insights: \textit{1)} Models with reflection mechanism demonstrate a superior CoT quality, with Kimi k1.5 outperforming GPT-4o and demonstrating the highest quality results; \textit{2)} CoT prompting often degrades LMM performance on perception-heavy tasks, suggesting a potentially harmful overthinking behavior; and \textit{3)} Although the CoT quality is high, LMMs with reflection exhibit significant inefficiency in both normal response and self-correction phases.
We hope MME-CoT serves as a foundation for advancing multimodal reasoning in LMMs.
\end{abstract}
\section{Introduction}
\label{submission}

\begin{figure}[t!]
    \centering
    \includegraphics[width=\linewidth]{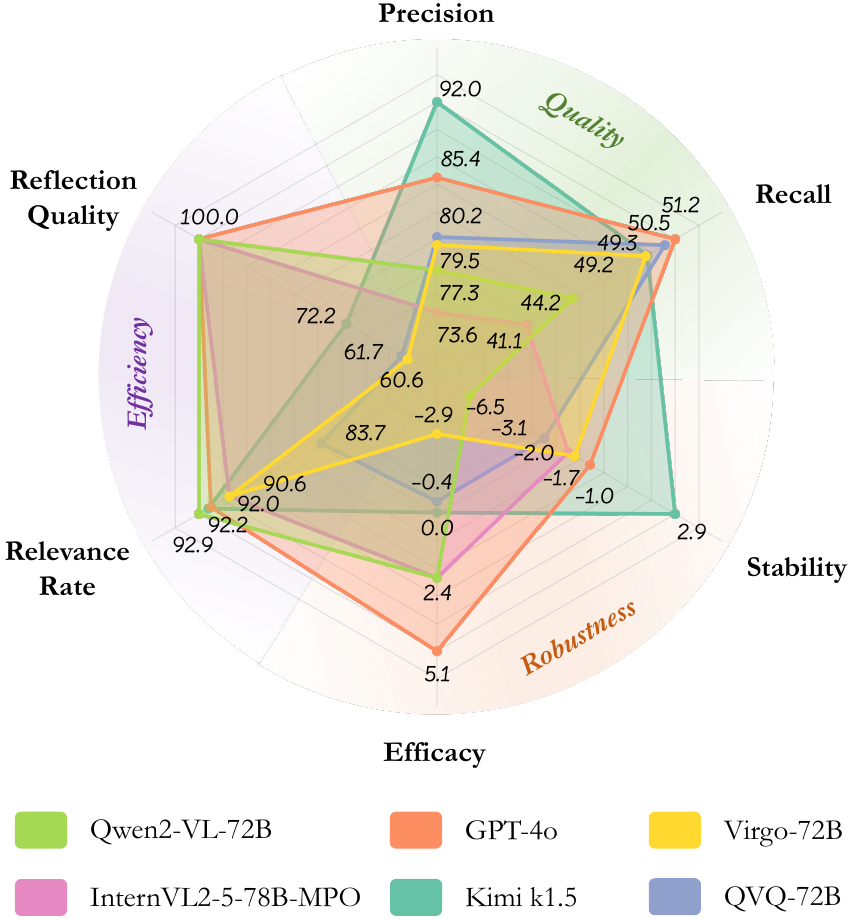}
    \caption{\textbf{Chain-of-Thought Performance of Leading LMMs in MME-CoT.} Our evaluation suite assesses LMMs using three novel metrics that yield six distinct scores. Results reveal that current open-source models, including those with reflection capabilities, still lag behind closed-source models like GPT-4o and Kimi k1.5 in key aspects of chain-of-thought reasoning.}
    \label{fig:teaser}
\end{figure}

\begin{figure*}[!t]
\centering
\includegraphics[width=\textwidth]{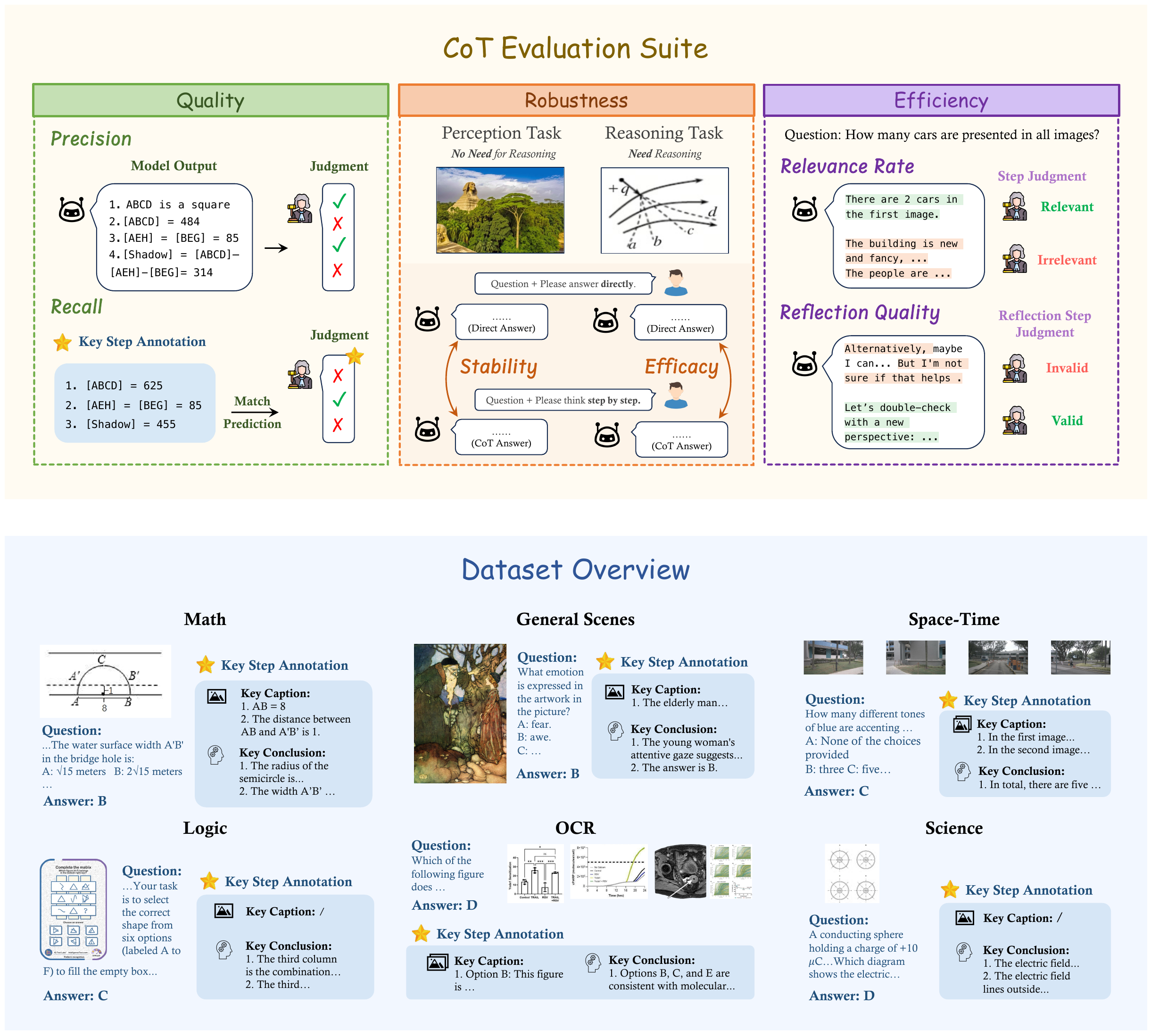} 
\caption{\textbf{An Overview of MME-CoT.} Our benchmark contains a comprehensive CoT evaluation suite with three novel aspects and a meticulously curated dataset encompassing six categories.}
\label{fig:demo1}
\vspace{-0.17cm}
\end{figure*}

The emergence of Chain-of-Thought (CoT)~\cite{wei2022chain} in Large Language Models (LLMs) has demonstrated promising advances in reasoning capabilities, exemplified by the recent OpenAI o1~\cite{o1} and DeepSeek-R1~\cite{guo2025deepseek}. By engaging in a more deliberate, stepwise reasoning process before reaching a final answer, this methodology presents an effective solution in tackling complex scenarios.

In parallel, the multimodal extensions of LLMs, termed Large Multimodal Models (LMMs), have demonstrated remarkable proficiency across diverse visual domains, e.g., general image recognition~\cite{zhang2023llava,zhu2023minigpt, openai2023gpt4v,zhang2024llama}, temporal video understanding~\cite{li2023videochat,chen2023videollm}, and 3D geometry perception~\cite{guo2024sam2point, xu2023pointllm,guo2023point,jia2024lift3d}. However, to what extent and how much CoT reasoning can benefit multimodal challenges still remains an open question. Although some previous efforts~\cite{zhang2024mathverse, yu2023mm,zhang2024mavis,guo2025can} have been made to evaluate the CoT capabilities of LMMs, their examination is insufficiently systematic and thorough, limiting our understanding of multimodal reasoning and its further development.

To bridge this gap, we propose \textbf{MME-CoT}, a comprehensive and specialized benchmark for evaluating the CoT reasoning skills within LMMs (Figure~\ref{fig:demo1}). Our benchmark spans six fundamental domains: math, science, OCR, logic, space-time, and general scenes, encompassing a broad range of CoT-relevant scenarios.
Unlike the simplistic metrics used in previous studies, MME-CoT introduces a rigorous evaluation framework that delves into the fine-grained CoT process of LMMs, assessing reasoning quality, robustness, and efficiency. Specifically, we address three critical research questions as follows:

\begin{enumerate}
    \item \textit{\textbf{Is each intermediate CoT step logically valid and faithful without hallucination?}}
    The outcome-oriented evaluation paradigm, where most current benchmark adapts, omits the scenario where the model reaches the correct answer through flawed logic or random guess. This causes an illusion of inflated reasoning capabilities in the model. To delve into the reasoning process, we introduce two interpretable metrics to evaluate \textbf{the Quality of CoT}: \textit{1) Recall}, which quantifies reasoning informativeness by measuring the proportion of ground-truth solution steps appearing in the response; \textit{2) Precision}, which measures faithfulness by evaluating how many of the generated steps are accurate. 

    \item \textit{\textbf{Does CoT interfere with perception tasks, and to what extent does it enhance reasoning tasks?}}
    While existing studies primarily focus on the performance improvements CoT brings to reasoning tasks, they often overlook whether CoT could inadvertently disrupt the model’s ability to solve perception tasks that require minimal reasoning.
    To this end, we present \textit{the first} investigation into \textbf{the Robustness of CoT} in LMMs. Our benchmark incorporates two task categories (perception and reasoning), and employs two distinct prompting strategies (`direct answer' and `step-by-step') to assess two metrics:
    \textit{1) Stability}, which examines whether CoT negatively impacts the model’s performance on direct perception tasks;
    \textit{2) Efficacy}, which measures the extent to which CoT enhances the model’s performance on complex reasoning tasks.

    \item \textit{\textbf{How can we assess the efficiency of CoT in a long reasoning process?}}
    Recent o1-like models have distinguished themselves by employing excessively long CoT and reflection steps. This raises a critical trade-off question: does this approach strike an optimal balance between accuracy and computational cost?
    To investigate this, we present \textit{the first} study on \textbf{the Efficiency of CoT} in LMMs. We evaluate efficiency using two key metrics:
    \textit{1) Relevance Rate}, which assesses the proportion of generated content that contributes to answering the question.
    \textit{2) Reflection Quality}, which analyzes whether each reflection step drives the question towards correctness.

\end{enumerate}

\begin{figure*}[!t]
\vspace{0.15cm}
\centering
\begin{minipage}[c]{0.63\textwidth}
\includegraphics[width=\columnwidth]{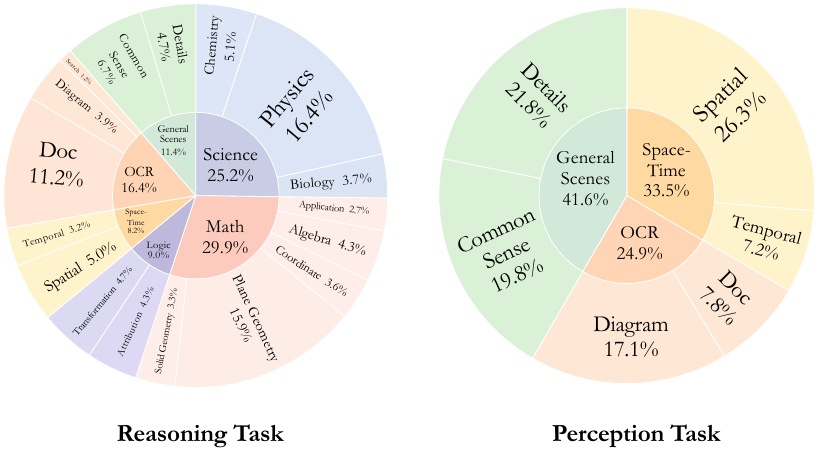}
\figcaption{\textbf{Category and Subcategory Distribution of MME-CoT.}} 
\label{pie}
\end{minipage} \hspace{3pt}   
\begin{minipage}[c]{0.34\textwidth}
\small
\centering
\vspace{-2pt}
\begin{adjustbox}{width=0.82\textwidth}
\begin{tabular}{lc}
\toprule
\textbf{Statistic} & \textbf{Number} \\
\midrule
Total questions & 1,130 \\
\quad - Reasoning questions & 837 (74.1\%) \\
\quad \quad Multiple-choice questions & 431 \\
\quad \quad Free-form questions & 406 \\
\quad - Perception questions & 293 (25.9\%) \\
\quad \quad Multiple-choice questions & 275 \\
\quad \quad Free-form questions & 18 \\
\midrule
Total key step annotation & 3,865 \\
\quad - Total inference conclusions & 2,667 \\
\quad - Average inference conclusions & 3.2 \\
\quad - Total image captions & 1,198 \\
\quad - Average image captions & 1.4 \\
Reference image caption item & 1,579 \\
Average reference caption & 1.9 \\
\midrule
Number of unique images & 2,380 \\
Number of unique questions & 808 \\
Number of unique answers & 271 \\
\midrule
Maximum question length & 477 \\
Maximum answer length & 15 \\
Average question length & 41.2 \\
Average answer length & 1.2 \\
\bottomrule
\end{tabular}
\end{adjustbox}
\vspace{-3pt}
\tabcaption{\textbf{Key Statistics of \dataset}.}
\label{table:statistics}
\end{minipage}
\vspace{-0.2cm}
\end{figure*}

Through our systematic evaluation and analysis, we discover that the fine-grained reflection capability greatly enhances the CoT quality, e.g., QVQ achieves F1 Score of 62.0\%, largely surpassing Qwen2-VL-72B by 6.8\%. Kimi k1.5 beats GPT-4o and achieves the best quality. As for the robustness, we surprisingly find that most models are interfered with by CoT on the perception tasks, implying a harmful overthinking behavior. The worst case happens in InternVL2.5-8B, where we witness a 6.8\% degradation when applying CoT on the perception tasks. This significantly impedes the applicability of models using CoT reasoning as a default practice. Moreover, for CoT efficiency, we notice that not all steps within the long CoT are related to answering the question, and the model could be distracted by the image content, especially when handling general scenes, space-time, and OCR tasks. Around 30\% to 40\% of reflection steps fail to help answer questions, pointing out critical issues of current models' reflection capabilities.

The contributions of this paper are summarized as follows:
\begin{itemize}
    \item The MME-CoT benchmark is curated, covering a comprehensive scope of six multimodal reasoning scenarios.
    The data collection and annotation process undergoes rigorous human verification, aiming to provide the community with a high-quality evaluation dataset for multimodal reasoning.

    \item We identify critical issues in existing benchmarks, and introduce a thorough evaluation suite specialized for multimodal CoT reasoning, which meticulously examines the reasoning quality, robustness, and efficiency.

    \item We conduct extensive experiments and analysis on state-of-the-art LMMs with reasoning capabilities. We summarize our observations and insights, hoping to inspire future advancements of reasoning performance.
\end{itemize}
\section{Dataset Curation}
\label{sec2_benchmark}

\subsection{Data Composition and Categorization.}
MME-CoT composes 6 major domains with 17 subcategories, as visualized in Fig.~\ref{pie}. Different from textual reasoning questions, the extra visual input significantly enriches the scope of the visual reasoning questions. 
With the image input, the model needs to frequently visit the image for relevant information according to current reasoning progress. Describing the image area of interest becomes a crucial part of the CoT process.
Thus, in addition to complex problems demanding rigorous logic, commonsense scenarios also pose a challenging reasoning problem,
as shown in the general scenes in Fig.~\ref{fig:demo1}. To maintain focus on the reasoning process, we exclude questions that require complex domain-specific theorems or specialized knowledge.

In addition, to evaluate CoT robustness detailed in Section~\ref{sec2_evaluation_robustness}, we incorporate a variety of perception tasks along with the reasoning tasks in the benchmark. 
The reasoning tasks contain questions that demand multi-step logical inference, while the perception tasks consist of questions that primarily test visual recognition abilities or require very minimal reasoning.
Existing benchmarks often conflate these two types of tasks, with perception and reasoning questions frequently appearing within the same categories. To address this, we implement a two-stage classification approach combining both model-based and human assessment. Initially, we leverage LMMs to guide the preliminary categorization by comparing their performance with and without CoT prompting. We employ GPT-4o~\cite{openai2024gpt4o} and Qwen2-VL-7B~\cite{wang2024qwen2} to answer questions using both direct and CoT approaches. Superior performance with CoT indicates a reasoning-dominant subcategory, while comparable or inferior CoT performance suggests either perception-focused content or insufficient model reasoning capabilities. The results are shown in Appendix~\ref{appendix:preliminary_result}. Subsequently, expert annotators review individual questions to finalize their classification. 
In total, MME-CoT contains 1,130 questions with 3,865 key step annotation. The detailed statistics of data compositions are shown in Table~\ref{table:statistics}. 
Please refer to Appendix~\ref{appendix:more_dataset} for more details about the distribution of data sources.

\subsection{Data Annotation and Review}
To facilitate CoT evaluation, we provide key steps annotation and reference image captions for all the reasoning questions. Key steps are defined as those that must be done to reach the correct answer. For efficient annotation, we first employ GPT-4o to generate the answer rationale and image captions. For the rationale, we provide both questions and ground truth answers to the model, which yields more accurate rationales compared to question-only prompting. Annotators are then asked to provide key intermediate steps with the help of GPT-4o's responses. For cases where GPT-4o fails to generate reasonable rationales, annotators develop solutions independently. The intermediate steps fall into two categories: inference conclusion and image caption. Note that the final answer is also included as a concluding inference. All the steps are reduced to the simplest form, retaining only core conclusions and relevant visual element descriptions. Notably, for problems with multiple solutions, annotators are required to provide all possible methods. For reference captions, we also ask annotators to verify and correct the details.

\begin{figure*}[!t]
\centering
\includegraphics[width=\textwidth]{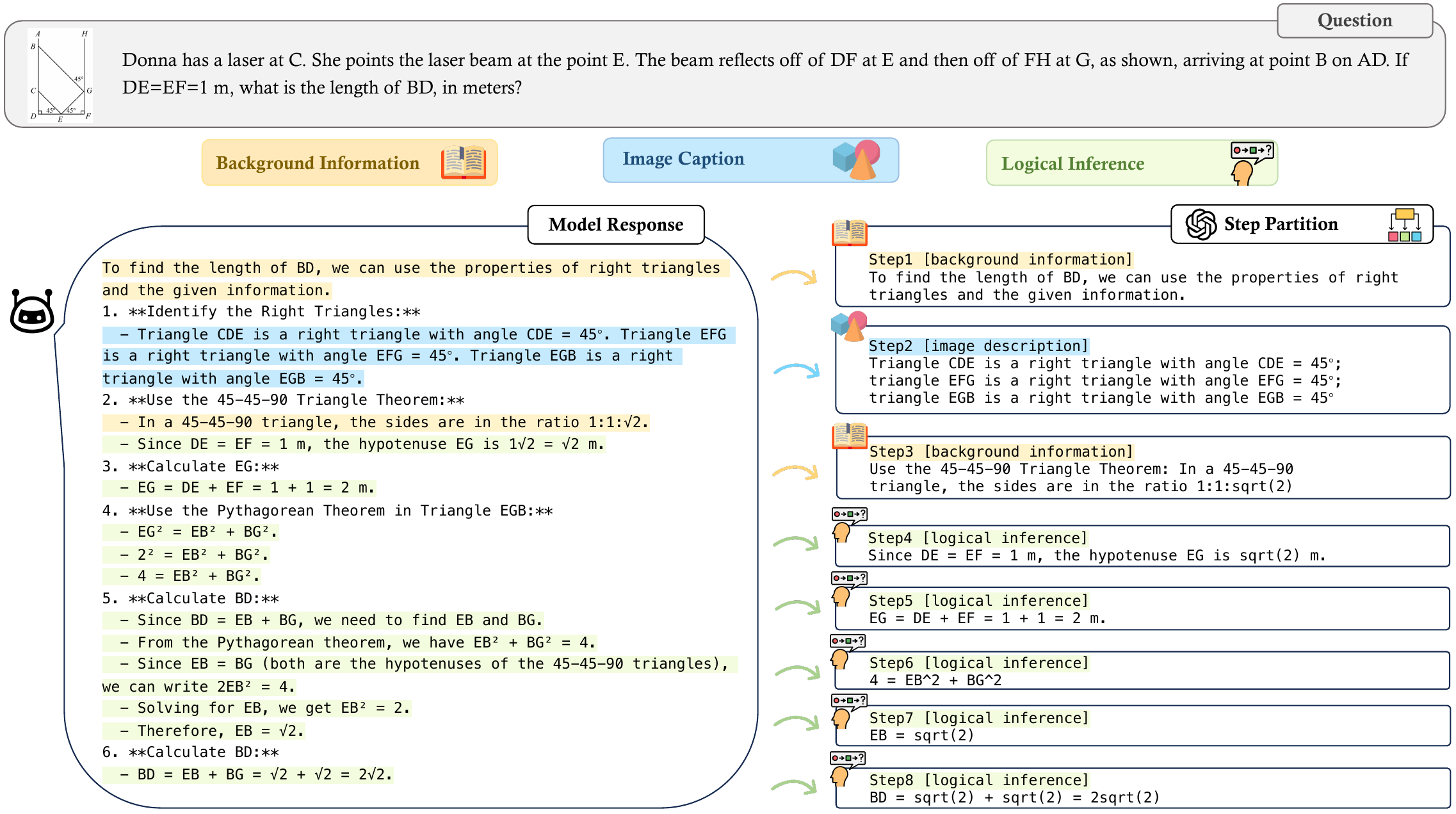} 
\caption{\textbf{Illustration of Step Partition.} We instruct GPT-4o to divide each step into three categories: image caption, background information, or logical inference. The step partition result is later used to perform step-wise reasoning evaluation. We focus on evaluating the image caption and logical inference steps, which are the keys to visual reasoning.}
\label{fig-step-partition}
\end{figure*}

\section{CoT Evaluation Strategy}
\label{sec2_evaluation}
Existing benchmarks only focus on evaluating the final answer of the questions, leaving the whole chain of thoughts unvisited. We argue that the CoT process reflects reasoning capability from multiple aspects, serving as a crucial medium to understand LMM's thinking pattern and deficiency. Here, we present the first holistic CoT evaluation suite to facilitate a comprehensive understanding of the LMMs' reasoning abilities. We detail the evaluation of correctness in Section~\ref{sec2_evaluation_correctness}, stability and efficacy in Section~\ref{sec2_evaluation_robustness}, and reflection quality in Section~\ref{sec2_evaluation_reflection}.

\subsection{CoT Quality Evaluation}
\label{sec2_evaluation_correctness}
\begin{figure*}[!t]
\centering
\includegraphics[width=\textwidth]{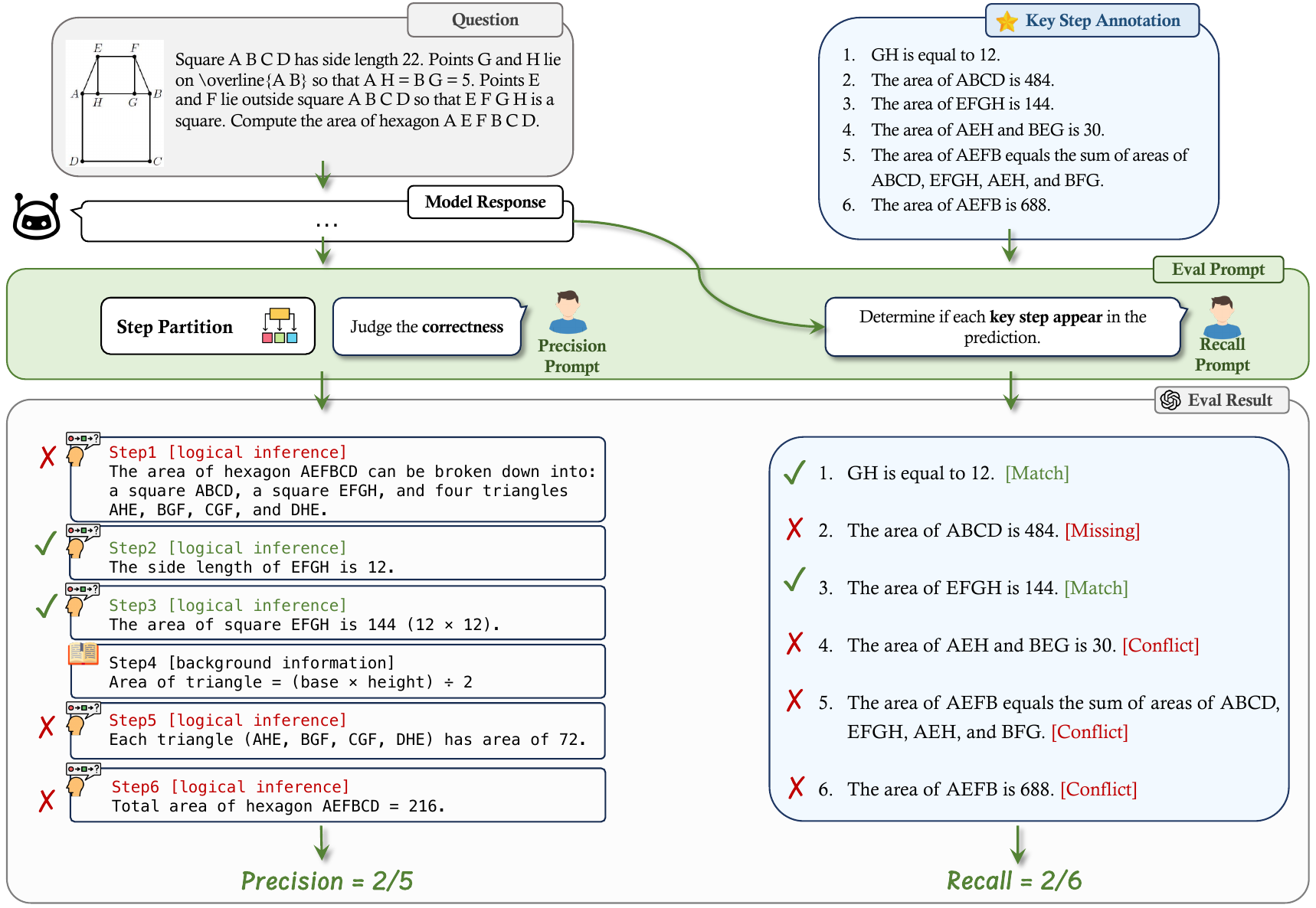} 
\caption{\textbf{Illustration of CoT Quality Evaluation.} For recall, we prompt GPT-4o to match each key step annotation in the prediction. For precision, GPT-4o is instructed to split the prediction into steps and determine the correctness of all the image caption and logical inference steps.}
\label{fig:quality}
\vspace{-0.17cm}
\end{figure*}

 Existing methods typically rely on state-of-the-art LLMs or LMMs to directly evaluate Chain-of-Thought reasoning based on self-defined criteria, using only the final answer as a reference~\cite{hao2024llm,zhang2024mathverse}. We identify two primary issues with the strategy. First, the scoring process only attends to the logical validity of each step, omitting the helpfulness evaluation. Second, there is a large number of complex visual reasoning questions that even the scoring model cannot solve. It is unreasonable for the scoring model to judge another model's reasoning process on these questions without knowing the ground truth solution process. 
 Therefore, building upon our annotated key steps and reference image captions, we leverage two interpretable metrics to evaluate the CoT correctness: recall and precision (Figure~\ref{fig:quality}). The two metrics respectively attend to the two aspects of the CoT correctness: informativeness and accuracy. We denote the key steps as $\mathcal{S} = \mathcal{C} \cup \mathcal{I}$, where $\mathcal{C} = \{c_1, ..., c_M\}$ includes $M$ key inference conclusions and $\mathcal{I} = \{i_1, ..., i_N\}$ includes $N$ key image captions.

\vspace{-1em}
 \paragraph{Recall.}
 We prompt GPT-4o~\cite{gpt4omini} to determine whether each key step occurs in the model's CoT response. Then we calculate the ratio of the matched key steps $\mathcal{S}_{\text{matched}}$ against all the annotated key steps:
\begin{align}
    &k_0 = \argmax_{k} \frac{\left | \mathcal{S}^{k}_{\text{matched}} \right |}{\left | \mathcal{S}^{k} \right |}, \\
    \text{Recall}_{\mathcal{C}} &= \frac{\left| \mathcal{C}^{k_0}_{\text{matched}} \right|}{\left| \mathcal{C}^{k_0} \right|}, \quad
    \text{Recall}_{\mathcal{I}} = \frac{\left| \mathcal{I}^{k_0}_{\text{matched}} \right|}{\left| \mathcal{I}^{k_0}\right|}, \\
    &\text{Recall} = \frac{\left | \mathcal{S}^{k_0}_{\text{matched}} \right |}{\left | \mathcal{S}^{k_0} \right |}.
\end{align}
 where $\mathcal{S}^{k}$ denotes the $k^{\text{th}}$ method of the problem.
 Intuitively, recall measures how many informative steps are reached by the model. From another perspective, this metric also strictly examines the process's rigorousness toward reaching the correct answer, eliminating the probability of random guessing. For questions with multiple methods, we compute the recall on the most matched method.

 \paragraph{Precision.} 
  We first instruct GPT-4o to partition the prediction into a sequence of steps $\mathcal{P}$, as shown in Fig.~\ref{fig-step-partition}.
  Each step is categorized into one of three classes: logical inference, image caption, and background information. The logical inference step draws an intermediate or final conclusion based on the previously obtained information. The image caption step depicts elements of interest in the image. The background information step states external knowledge or question information. Visual reasoning can be primarily characterized as an interleaved sequence of image captions and logical inferences, so we focus on measuring precision for these two key step types. We assess the correctness of logical inference steps ($\mathcal{C}^{\mathcal{P}}$) and image caption steps ($\mathcal{I}^{\mathcal{P}}$) using two criteria: 1. If the step exists in $\mathcal{S}$, the step is correct. 2. If the step is logically correct or faithfully depicts the image based on the annotations, the step is also correct. Thus, we compute precision as:
\begin{align}
    \text{Precision}_{\mathcal{C}} &= \frac{\left| \mathcal{C}^{\mathcal{P}}_{\text{correct}} \right|}{\left| \mathcal{C}^{\mathcal{P}} \right|}, \quad
    \text{Precision}_{\mathcal{I}} = \frac{\left| \mathcal{I}^{\mathcal{P}}_{\text{correct}} \right|}{\left| \mathcal{I}^{\mathcal{P}} \right|}, \\
    & \text{Precision} = \frac{\left| \mathcal{C}^{\mathcal{P}}_{\text{correct}} \cup \mathcal{I}^{\mathcal{P}}_{\text{correct}} \right|}{\left| \mathcal{C}^{\mathcal{P}} \cup \mathcal{I}^{\mathcal{P}} \right|}
\end{align}
Intuitively, precision evaluates the faithfulness of each step, considering all the possible reasoning output. Finally, we calculate the F1 score as the metric of CoT quality.

\subsection{CoT Robustness Evaluation}
\label{sec2_evaluation_robustness}
\begin{figure}[t]
\begin{center}
\centerline{\includegraphics[width=\columnwidth]{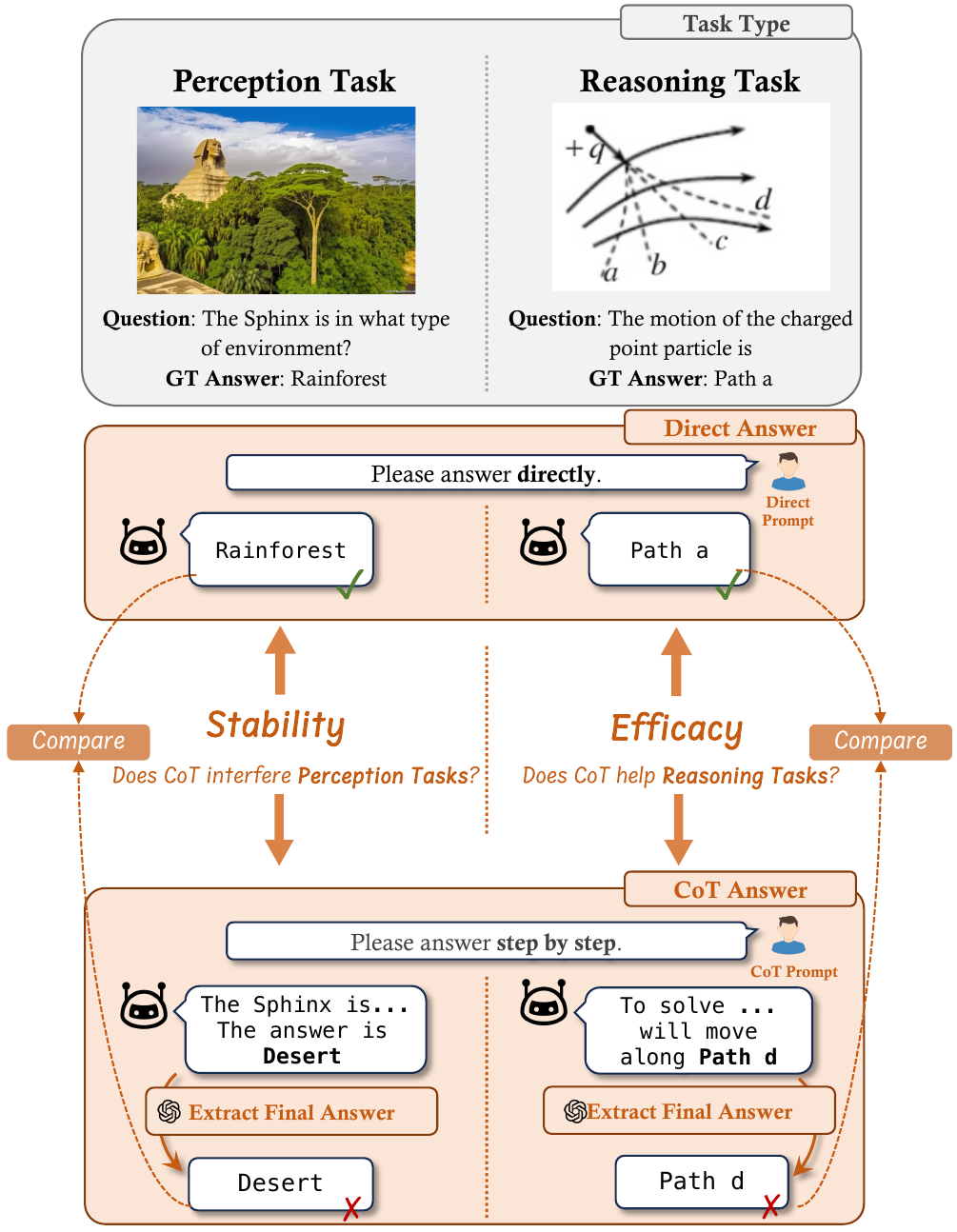}}
\caption{\textbf{Illustration of CoT Robustness Evaluation.} We compare the performance of applying CoT prompt and direct prompt on two types of tasks: perception and reasoning. The stability score measures whether CoT interferes with perception, while the efficacy score assesses the performance gain of CoT on reasoning tasks.}
\label{fig:robustness}
\end{center}
\vskip -0.2in
\end{figure}

Here, we perform the first investigation on the robustness of CoT in visual reasoning. The effectiveness of CoT on reasoning tasks has been verified in many works~\cite{wei2022chain, o1}. However, how CoT impacts visual perception tasks or tasks requiring minimal reasoning still remains unknown. Despite the neglect, this question bears great importance.
In real-world applications, what task is given is unknown in advance. Whether the model should perform CoT to solve the task is difficult to determine. In fact, there exists no golden standard to determine which question can benefit from CoT so far~\cite{sprague2024cot}.
Instead of trying to define this criterion, we examine the performance of CoT across all kinds of tasks, both reasoning and perception. We argue that an ideal CoT process should assist in reasoning and not interfere with pure perception. Therefore, it can be applied for any tasks.
Based on this, we propose to evaluate two metrics of CoT: stability and efficacy (Figure~\ref{fig:robustness}). We leverage two kinds of prompts: the direct prompt ($\textsc{dir}$) and the CoT prompt ($\textsc{cot}$). The direct prompt asks the model to directly provide the final answer, while the CoT prompt instructs the model to perform step-by-step reasoning and finally give the answer. To directly compare the performance difference caused by these two prompts, we conduct the direct evaluation, which only judges the correctness of the final answer, i.e., accuracy. We instruct GPT-4o mini~\cite{gpt4omini} to extract the final answer, and then compare it with the ground truth answer, following the two-step procedure introduced in~\cite{zhang2024mathverse}. 

\begin{figure}[t]
\begin{center}
\centerline{\includegraphics[width=\columnwidth]{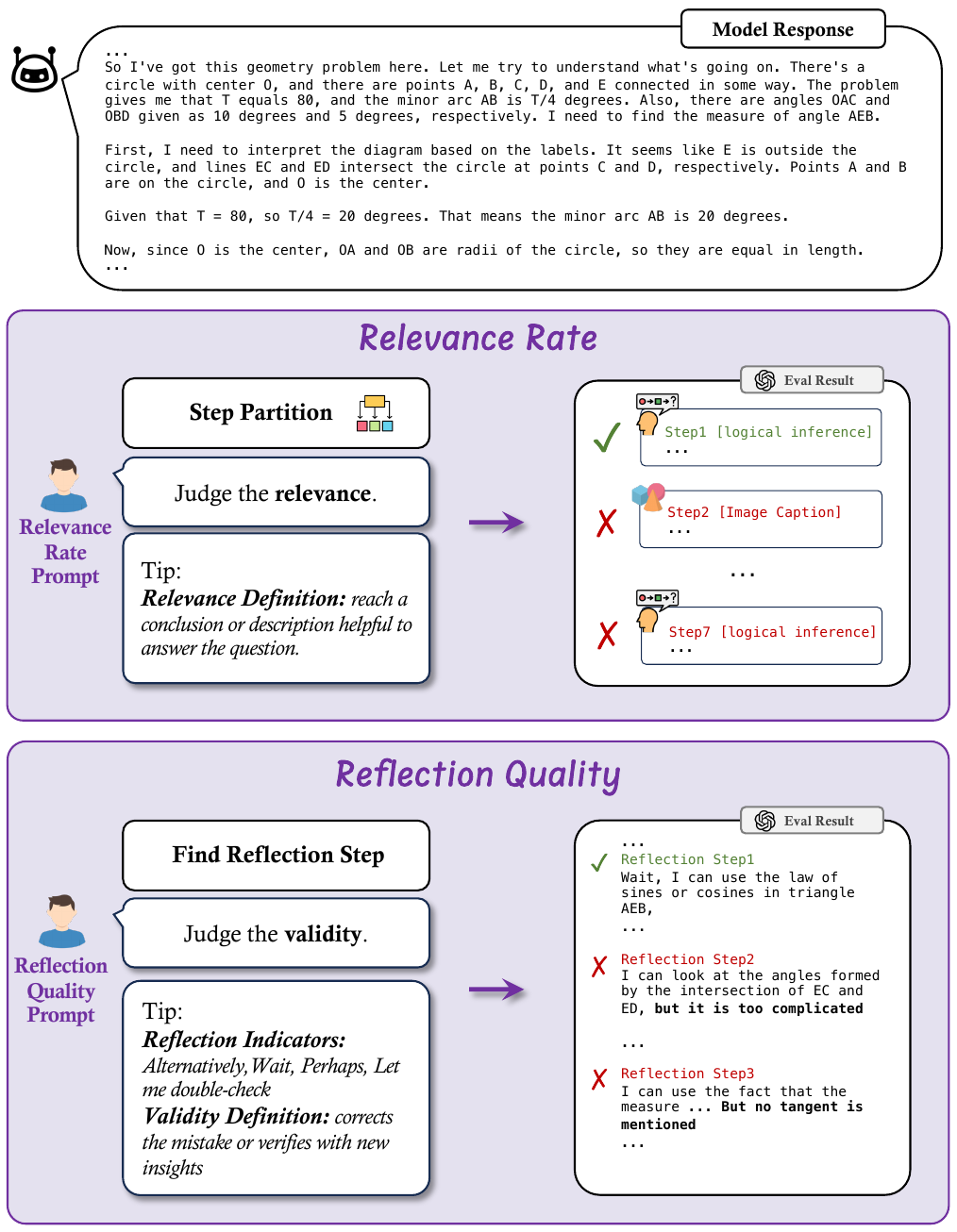}}
\caption{\textbf{Illustration of CoT Efficiency Evaluation.} For relevance rate, we partition the prediction into steps and determine if it is relevant by GPT-4o. For reflection quality, we prompt GPT-4o to identify the reflection steps by common indicators and judge the validity of the reflection. The definitions of relevance and validity are included.}
\label{fig:efficiency}
\end{center}
\vskip -0.2in
\end{figure}

\vspace{-1em}
\paragraph{Stability.}
We define the performance difference of the two prompts on the perception tasks $\mathbf{P}$ as the stability score:
\begin{equation}
    \text{Stability} = \text{Acc}^{\mathbf{P}}_{\textsc{cot}} - \text{Acc}^{\mathbf{P}}_{\textsc{dir}}.
\end{equation}
Intuitively, applying the CoT prompt to perception tasks should not degrade performance compared with the direct prompt. Thus, a model with stable CoT should be not less than 0. Otherwise, the model's thinking process demonstrates inconsistency and harm. The overthinking process pushes over the original correct judgment.

\vspace{-1em}
\paragraph{Efficacy.}
Similarly, the performance difference of the two prompts on the reasoning tasks $\mathbf{R}$ is defined as the score:
\begin{equation}
    \text{Efficacy} = \text{Acc}^{\mathbf{R}}_{\textsc{cot}} - \text{Acc}^{\mathbf{R}}_{\textsc{dir}}.
\end{equation}
Intuitively, CoT facilitates stepwise thinking and therefore benefits answering reasoning tasks. The difference reflects how much CoT can enhance reasoning.

\begin{table*}[!t]
\centering
\caption{\textbf{Evaluation Results of Three Aspects of CoT in \dataset.} We mark the highest score of each metric in \colorbox{backred!60}{red}. $*$ denotes unreliable results due to the refusal to answer directly.}
\vspace{-3pt}
\renewcommand\tabcolsep{2.5pt}
\renewcommand\arraystretch{1.25}
\resizebox{1.0\linewidth}{!}{
\begin{tabular}{l|>{\columncolor{backred!20}}c|>{\columncolor{lightgray}}c|cc|>{\columncolor{lightgray}}c|cc|>{\columncolor{backred!20}}c|>{\columncolor{lightgray}}c|cc|>{\columncolor{lightgray}}c|cc|>{\columncolor{backred!20}}c|>{\columncolor{lightgray}}c|cc|>{\columncolor{lightgray}}c}
\toprule
\multirow{3}*{\makecell*[l]{\large Model}}
&\multicolumn{7}{c|}{\makecell*[c]{CoT Quality}}
&\multicolumn{7}{c|}{CoT Robustness}
&\multicolumn{5}{c}{CoT Efficiency}\\
\cmidrule{2-20}
& \makecell{F1\\Score} & \makecell{Precision} & \makecell{Image} & \makecell{Conclusion} & \makecell{Recall} & \makecell{Image} & \makecell{Conclusion} & \makecell{Avg.\\Score} & \makecell{Stability} & \makecell{CoT\\Perception} & \makecell{Direct\\Perception} & \makecell{Efficacy} & \makecell{CoT\\Reasoning} & \makecell{Direct\\Reasoning} & \makecell{Avg.\\Score} & \makecell{Relevance\\Rate} & \makecell{Image} & \makecell{Conclusion} & \makecell{Reflection\\Quality} \\
\midrule
\cmidrule{1-20}
\multicolumn{20}{c}{\textit{Open-source LMMs}}\\
\cmidrule{1-20}
Mulberry & 27.4 & 59.1 & 74.1 & 53.8 & 17.8 & 26.5 & 17.1 & \colorbox{backred!60}{3.5*} & \colorbox{backred!60}{4.4*} & 42.3 & 37.9 & 2.6* & 18.6 & 16.0 & 89.5 & 79.0 & 50.8 & 95.4 & 100 \\
LLaVA-OV-7B & 30.9 & 50.9 & 47.2 & 43.5 & 22.2 & 24.4 & 23.2 & -3.4 & -3.8 & 46.1 & 49.8 & -3.0 & 16.4 & 19.4 & 91.5 & 83.0 & 72.1 & 93.6 & 100 \\
LLaVA-CoT & 34.9 & 53.9 & 75.6 & 46.2 & 25.8 & 35.8 & 24.4 & 0.4* & 1.4* & 51.5 & 50.2 & -0.6* & 24.4 & 25.0 & 94.0 & 88.1 & 69.2 & 96.2 & 100 \\
LLaVA-OV-72B & 36.3 & 57.3 & 43.4 & 50.6 & 26.6 & 29.5 & 27.4 & -0.2 & 0.3 & 61.1 & 60.8 & -0.6 & 27.6 & 28.2 & 95.4 & 90.8 & 83.7 & 98.3 & 100 \\
MiniCPM-V-2.6 & 39.8 & 57.3 & 63.4 & 45.4 & 30.5 & 47.5 & 26.7 & -3.5 & -4.8 & 59.4 & 64.2 & -2.2 & 26.2 & 28.3 & 92.8 & 85.7 & 74.6 & 97.6 & 100 \\
InternVL2.5-8B & 41.1 & 60.0 & 52.4 & 50.8 & 31.3 & 40.4 & 30.6 & -3.0 & -6.8 & 57.3 & 64.2 & 0.9 & 30.3 & 29.4 & \colorbox{backred!60}{98.4} & \colorbox{backred!60}{96.8} & \colorbox{backred!60}{93.0} & 98.9 & 100 \\
Qwen2-VL-7B & 42.1 & 61.6 & 61.0 & 49.3 & 32.0 & 46.6 & 30.5 & -4.0 & -3.1 & 60.1 & 63.1 & -4.8 & 26.0 & 30.8 & 94.9 & 89.8 & 80.3 & 98.8 & 100 \\
InternVL2.5-8B-MPO & 43.0 & 60.4 & 60.8 & 49.9 & 33.4 & 44.9 & 31.8 & 0.6 & 0.3 & 62.5 & 62.1 & 0.9 & 28.8 & 27.9 & 94.7 & 89.3 & 84.0 & 96.4 & 100 \\
InternVL2.5-78B-MPO & 52.7 & 73.6 & 68.4 & 63.0 & 41.1 & 53.6 & 39.1 & 0.2 & -2.0 & 68.3 & 70.3 & 2.4 & 38.0 & 35.6 & 95.3 & 90.6 & 82.9 & 98.2 & 100 \\
Qwen2-VL-72B & 56.2 & 77.3 & 67.2 & 70.3 & 44.2 & 57.1 & 42.2 & -2.1 & -6.5 & 68.9 & 75.4 & 2.4 & 38.6 & 36.2 & 96.5 & 92.9 & 86.0 & 98.7 & 100 \\
Virgo-72B & 60.8 & 79.5 & 71.6 & 72.7 & 49.2 & 60.5 & 47.7 & -2.3* & -1.7* & 74.1 & \colorbox{backred!60}{75.8} & -2.9* & \colorbox{backred!60}{41.8} & \colorbox{backred!60}{44.7} & 75.3 & 90.6 & 79.8 & 95.6 & 60.6 \\
QVQ-72B & 62.0 & 80.2 & 73.9 & 77.5 & 50.5 & 60.1 & 48.9 & -1.8* & -3.1* & \colorbox{backred!60}{69.6} & 72.7 & -0.4* & 41.0 & 41.3 & 67.9 & 83.7 & 63.9 & 95.1 & 61.7 \\
\midrule
\cmidrule{1-20}
\multicolumn{20}{c}{\textit{Closed-source LMMs}}\\
\cmidrule{1-20}
GPT-4o & 64.0 & 85.4 & 73.3 & 81.4 & \colorbox{backred!60}{51.2} & \colorbox{backred!60}{64.3} & \colorbox{backred!60}{49.9} & 2.1 & -1.0 & 71.0 & 72.0 & \colorbox{backred!60}{5.1} & 40.6 & 35.5 & 96.0 & 92.0 & 82.4 & \colorbox{backred!60}{99.1} & 100 \\
Kimi k1.5 & \colorbox{backred!60}{64.2} & \colorbox{backred!60}{92.0} & \colorbox{backred!60}{78.1} & \colorbox{backred!60}{89.8} & 49.3 & 62.9 & 47.9 & 1.4* & 2.9* & 65.7 & 62.9 & 0.0* & 40.0 & 40.0 & 82.2 & 92.2 & 82.2 & 97.2 & 72.2 \\
\bottomrule
\end{tabular}
}
\label{table:main_result}
\end{table*}

\subsection{CoT Efficiency Evaluation}
\label{sec2_evaluation_reflection}
Models like o1 generate extremely long thinking processes with reflection and verification of current steps and outcomes. We perform the first exhaustive analysis of the CoT efficiency of visual reasoning with two carefully designed metrics (Figure~\ref{fig:efficiency}):
\paragraph{Relevance Rate.}
Although the long reasoning content allows for deeper thinking, it may also introduce a large amount of irrelevant information.
As shown in the bottom left of Fig.~\ref{fig:efficiency}, the model has identified the critical element in the image for answering the question, but it still generates a detailed description of other objects.
This irrelevant information provides no helpful information to work out the answer. In the meantime, this extra content slows down the generation speed. Similar to the calculation of precision, we employ the same method to partition the prediction into steps. Then, we instruct GPT-4o to determine all the relevant steps $\mathcal{P}_\text{relevant}$. The step is considered relevant only when the majority of its content works towards solving the question. 
We first compute the raw relevance rate and then apply a scaling factor to amplify the differences between models. Let $r_x$ denote the raw relevance rate:
\begin{align}
r_{\mathcal{C}} = \frac{\left| \mathcal{C}^{\mathcal{P}}_{\text{relevant}} \right|}{\left| \mathcal{C}^{\mathcal{P}} \right|}, \quad
r_{\mathcal{I}} = \frac{\left| \mathcal{I}^{\mathcal{P}}_{\text{relevant}} \right|}{\left| \mathcal{I}^{\mathcal{P}} \right|},
\end{align}
\vspace{-0.6cm}
\begin{align}
r = \frac{\left| {\mathcal{P}}_{\text{relevant}} \right|}{\left| \mathcal{P} \right|}.
\end{align}

Then, the final relevance rate $\text{Relevance Rate}_{x}$ is defined as:
\begin{align}
\text{Relevance Rate}_{x} = \frac{r_x-\alpha}{1-\alpha}, \quad x \in {\mathcal{C}, \mathcal{I}, \emptyset}
\end{align}
where $x = \emptyset$ corresponds to the overall relevance rate, and we take $\alpha$ as $0.8$.

\vspace{-1em}
\paragraph{Reflection Quality.}
The superior reasoning ability could be largely attributed to the reflection and verification process. However, our analysis reveals that not all reflective steps contribute meaningfully to finding correct answers. We identify distinct failure patterns in the reflection process. Some reflective steps mislead the reasoning by introducing new errors or incorrect assumptions, while others are redundant, simply echoing previous conclusions without contributing new insights. To account for failure reflection scenarios, we propose to measure the validity of the reflection. We define a valid reflection as either correctly pointing out the previous mistakes or verifying the previous conclusion with a new insight. Otherwise, the reflection only slows down the reasoning.
To instruct GPT-4o to determine all the valid reflection steps $\mathcal{R}$, we list a set of common indicators of the start of the reflection, such as ``Wait" and ``Alternatively", and illustrate the definition of valid reflection. For all the valid reflection steps $\mathcal{R}_\text{valid}$, the reflection quality is computed as:
\begin{equation}
    \text{Reflection Quality} =  \frac{\left| {\mathcal{R}}_{\text{valid}} \right|}{\left| \mathcal{R} \right|}.
\end{equation}

\section{Experiments}
In this section, we conduct a systematic evaluation of state-of-the-art models on \dataset. We first detail the experiment setup in Section~\ref{sec:exp_steup}. Then in Section~\ref{sec:exp_quantitative}, we report the quantitative results and provide valuable insights derived from our analysis.

\subsection{Experiment Setup}
\label{sec:exp_steup}
\paragraph{Evaluation Models.} 
We select top-performing LMMs for comprehensive CoT evaluation. We test earlier models such as LLaVA-OneVision (7B, 72B)~\cite{li2024llava-ov}, Qwen2-VL (7B, 72B)~\cite{Qwen2-VL}, MiniCPM-V-2.6~\cite{yao2024minicpm}, and InternVL2.5 (8B)~\cite{chen2024expanding}, which are not trained for the reasoning capability. We also include GPT-4o~\cite{openai2024gpt4o} as a strong baseline model.
Besides, we test recent models targeting reasoning, including LLaVA-CoT (11B)~\cite{xu2024llavacot}, Mulberry (8B)~\cite{yao2024mulberry}, InternVL2.5-MPO (8B, 78B)~\cite{wang2024mpo}.
Finally, we evaluate LMMs with reflection capabilities, including both closed-source models like Kimi k1.5~\cite{team2025kimi} and open-source implementations such as QVQ-72B~\cite{qvq-72b-preview} and Virgo-72B~\cite{du2025virgo}.

Note that we sample 150 questions from \dataset to evaluate Kimi k1.5, due to the access limitations. The sample comprises 115 reasoning and 35 perception questions. 

\begin{table*}[!t]
\centering
\caption{\textbf{Evaluation Results of Three Aspects of CoT in Each Category in \dataset.} Best performance is marked in \colorbox{backred!60}{red}.  $*$ denotes unreliable results due to the refusal to answer directly.}
\vspace{-3pt}
\renewcommand\tabcolsep{2.0pt}
\renewcommand\arraystretch{1.25}
\resizebox{1.0\linewidth}{!}{
\begin{tabular}{l|ccc|ccc|ccc|cc|cc|cc}
\toprule
\multirow{2}*{\makecell*[l]{\large Model}} & \multicolumn{3}{c|}{\makecell*[c]{General Scenes}} & \multicolumn{3}{c|}{Space-Time} & \multicolumn{3}{c|}{OCR} & \multicolumn{2}{c|}{Math} & \multicolumn{2}{c|}{Science} & \multicolumn{2}{c}{Logic} \\
& Quality & Robustness & Efficiency & Quality & Robustness & Efficiency & Quality & Robustness & Efficiency & Quality & Efficiency & Quality & Efficiency & Quality & Efficiency \\
\midrule
Mulberry & 33.9 & \colorbox{backred!60}{4.3} & 76.0 & 18.2 & 1.0 & 38.4 & 26.7 & \colorbox{backred!60}{6.6} & 26.4 & 29.1 & 87.9 & 29.1 & 91.9 & 13.9 & \colorbox{backred!60}{99.1} \\
LLaVA-OV-7B & 41.8 & -6.2 & 81.8 & 23.8 & -6.7 & 24.8 & 44.1 & -0.2 & 42.7 & 27.4 & 97.3 & 28.5 & 95.1 & 12.2 & 98.0 \\
LLaVA-CoT & 38.2 & -2.2 & 89.9 & 33.6 & 2.8 & 68.9 & 37.4 & 0.0 & 77.8 & 35.3 & 91.0 & 36.4 & 93.4 & 14.9 & 97.1 \\
LLaVA-OV-72B & 41.8 & -2.3 & \colorbox{backred!60}{98.9} & 29.0 & -0.9 & 43.6 & 40.8 & -1.7 & 84.2 & 38.4 & 98.7 & 35.4 & 95.7 & 18.4 & 82.3 \\
MiniCPM-V-2.6 & 47.1 & 3.2 & 87.7 & 49.3 & -14.4 & 71.1 & 63.7 & -4.9 & 62.0 & 32.9 & 95.2 & 29.5 & 90.4 & 16.9 & 93.7 \\
InternVL2.5-8B & 43.8 & -6.4 & 87.1 & 50.7 & -8.9 & \colorbox{backred!60}{99.1} & 44.7 & -4.1 & \colorbox{backred!60}{98.9} & 40.9 & 98.0 & 40.8 & 97.1 & 19.5 & 96.8 \\
Qwen2-VL-7B & 46.7 & -3.4 & 79.3 & 51.7 & -11.8 & 73.0 & 65.9 & 0.9 & 86.2 & 34.0 & 97.9 & 34.6 & 95.0 & 18.4 & 76.7 \\
InternVL2.5-8B-MPO & 47.2 & 2.9 & 94.3 & 51.8 & -0.2 & 74.6 & 59.6 & -1.0 & 81.5 & 37.4 & 93.4 & 39.0 & 95.6 & 20.9 & 79.9 \\
InternVL2.5-78B-MPO & 47.9 & 0.0 & 89.3 & 55.5 & -2.3 & 91.9 & 72.2 & 2.2 & 73.1 & 50.6 & 95.1 & 48.5 & 97.7 & 24.2 & 87.2 \\
Qwen2-VL-72B & 51.9 & -2.9 & 88.9 & 59.7 & -5.3 & 86.7 & 77.6 & 2.5 & 81.7 & 49.6 & 97.8 & 53.6 & \colorbox{backred!60}{99.0} & 40.0 & 88.0 \\
Virgo-72B & 60.5 & 0.5 & 91.0 & 59.6 & -3.8 & 86.0 & 79.9 & -1.0 & 82.1 & 59.6 & 90.3 & 55.5 & 98.7 & 39.6 & 88.2 \\
QVQ-72B & \colorbox{backred!60}{62.6} & -1.5 & 86.9 & 58.2 & -2.5 & 57.7 & 76.9 & -1.4 & 52.6 & \colorbox{backred!60}{61.4} & 92.7 & 57.7 & 95.9 & \colorbox{backred!60}{44.6} & 94.9 \\
GPT4o & 62.3 & -1.7 & 96.2 & \colorbox{backred!60}{66.3} & \colorbox{backred!60}{5.5} & 64.7 & \colorbox{backred!60}{83.3} & -1.0 & 82.1 & 60.8 & \colorbox{backred!60}{98.8} & \colorbox{backred!60}{64.1} & 97.4 & 27.2 & 92.0 \\
\bottomrule
\end{tabular}
}
\label{table:category_result}
\end{table*}

\paragraph{Implementation Details.}
We define the CoT prompt as: \textit{Please generate a step-by-step answer, include all your intermediate reasoning process, and provide the final answer at the end.} and the direct prompt as: \textit{Please directly provide the final answer without any other output.}
We only calculate recall of image observation and logical inference on questions where key inference conclusion or image observation exists.
We employ GPT-4o mini for the direct evaluation and GPT-4o for all other criteria. For hyperparameters, we follow the settings in VLMEvalKit~\cite{duan2024vlmevalkit}. 

\subsection{Quantitative Results}
\label{sec:exp_quantitative}
We conduct extensive experiments on various LMMs with our proposed CoT evaluation suite. 
The main results are presented in Table~\ref{table:main_result} and Table~\ref{table:category_result}. We begin by analyzing the overall performance and then highlight key findings.
\paragraph{Overall Results.}
In Table~\ref{table:main_result}, we present
the overall performance of three CoT evaluation perspectives with specific metrics. 
To provide a comprehensive understanding, we report precision, recall, and relevance for both logical inference and image caption steps. For robustness, we provide the direct evaluation result on the perception and reasoning tasks, with either CoT or direct prompt. We employ the average value of the stability and efficacy as the final robustness metric. Notably, we define the reflection quality as 100 on models incapable of reflection.

For CoT quality, Kimi k1.5 achieves the highest F1 score. Open-source models with larger sizes consistently demonstrate better performance, highlighting the scalability of LMMs. Notably, Qwen2-VL-72B outperforms all other open-source models without reflection, even surpassing InternVL2.5-78B-MPO, which is specifically enhanced for reasoning. Analysis reveals that GPT-4o achieves superior performance across all recall metrics, while Kimi k1.5 demonstrates the highest scores in precision evaluations.
For CoT robustness, Mulberry obtains the highest average score. However, when we look into its output, we find it still generates lengthy rationales despite receiving a direct prompt. Even worse, the direct prompt seems to be an out-of-distribution input for Mulberry, 
frequently leading to nonsensical outputs. Further analysis of other models’ predictions reveals that LLaVA-CoT, Virgo, QVQ, and Kimi k1.5 similarly neglect the direct prompt, instead generating extended rationales before answering. Consequently, their robustness scores may be misleading. Once again, GPT-4o achieves the highest robustness score. Among open-source models, only InternVL2.5-MPO, in both its 8B and 78B variants, attains a positive robustness score.
Finally, for CoT efficiency, InternVL2.5-8B obtains the maximum relevance of 98.4\%, suggesting its consistent focus on questions.

Now, we summarize our key observations as follows:
\paragraph{\textit{Models with reflection largely benefit CoT quality.}}
As shown in Table~\ref{table:main_result}, the F1 scores of the two models with reflection capability most closely approach GPT-4o. After specifically fine-tuning for the reasoning capabilities from Qwen2-VL-72B, QVQ surpasses its base model by 5.8\%. Notably, although QVQ generates longer CoT sequences than Qwen2-VL-72B, QVQ's precision still exceeds Qwen2-VL-72B by 2.9\%, indicating superior accuracy in each reasoning step. Kimi k1.5 also surpasses the previous state-of-the-art model GPT-4o, obtaining the highest CoT quality.

\paragraph{\textit{Long CoT does not necessarily cover key steps.}} 
Despite high precision in long CoT models, the informativeness of each step is not guaranteed. We observe that the recall trend among GPT-4o, QVQ, and Virgo does not align with their CoT Rea. performance (i.e., their final answer accuracy on the reasoning tasks under the CoT prompt). Specifically, while both Virgo and QVQ outperform GPT-4o in direct evaluation, they lag behind in recall. This suggests that long CoT models sometimes reach correct answers while skipping intermediate steps, which contradicts the principle of stepwise reasoning and warrants further investigation.

\paragraph{\textit{CoT impairs perception task performance in most models.}}
Surprisingly, most models exhibit negative stability scores, indicating that CoT interferes with perception tasks. The most significant degradation occurs in InternVL2.5-8B, where performance drops by 6.8\%. This reveals inconsistency and potential overthinking in current models, presenting a significant barrier to adopting CoT as the default answering strategy. Among models that provide direct answers, only LLaVA-OV-72B and InternVL2.5-8B-MPO achieve a modest positive score of 0.3\%.

\paragraph{\textit{More parameters enable models to grasp reasoning better.}} 
We find that models with larger parameter counts tend to achieve higher efficacy scores. This pattern is evident across LLaVA-OV, InternVL2.5-MPO, and Qwen2-VL. For instance, while Qwen2-VL-7B shows a 4.8\% decrease in performance when applying CoT to reasoning tasks, its larger counterpart, Qwen2-VL-72B, demonstrates a 2.4\% improvement. This discrepancy suggests that models with more parameters could better grasp the reasoning ability under the same training paradigm.

\paragraph{\textit{Long CoT models may be more susceptible to distraction.}} 
Long CoT models may demonstrate lower relevance scores compared to other models. They frequently generate content unrelated to solving the given question, corresponding to their relatively low recall scores compared to direct evaluation, like QVQ. Although a few models with short CoT, like Mulberry and LLaVA-OV-7B, also obtain a low relevance rate, we find that it is because these models may keep repeating words when dealing with specific type of questions, resulting in irrelevant judgment. The fine-grained metric reveals that models tend to lose focus when describing images, often producing exhaustive captions regardless of their relevance to the question. From Table~\ref{table:category_result}, we find that this phenomenon prevails in general scenes, space-time, and OCR tasks. This behavior can significantly slow inference by generating substantial irrelevant content. Teaching long CoT models to focus on question-critical elements represents a promising direction for future research.

\paragraph{\textit{Reflection often fails to help.}} 
While reflection is a key feature of long CoT models for answer verification, both QVQ and Virgo achieve reflection quality scores of only about 60\%, indicating that approximately 40\% of reflection attempts fail to contribute meaningfully to answer accuracy. Even for the closed-source model Kimi k1.5, over 25\% reflection steps are also invalid. This substantial failure rate compromises efficiency by potentially introducing unnecessary or distracting steps before reaching correct solutions. Future research should explore methods to reduce these ineffective reflections to improve both efficiency and quality.

\begin{figure}[t]
\begin{center}
\vspace{0.2cm}
\centerline{\includegraphics[width=0.8\columnwidth]{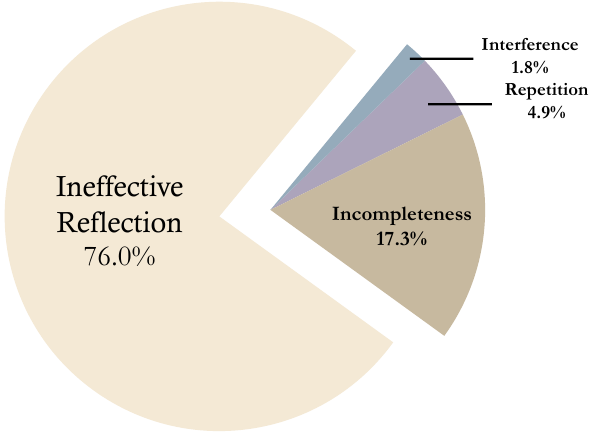}}
\caption{\textbf{Distribution of Reflection Error Types.} We identify four types of error: ineffective reflection, incompleteness, repetition, and interference.}
\label{fig:ref_error_distribution}
\end{center}
\vspace{-0.6cm}
\end{figure}

\subsection{Error Analysis}
\label{sec:exp_analysis}
In this section, we analyze error patterns in the LMM reflection process. An effective reflection should either correct previous mistakes or validate correct conclusions through new insights. We examined 200 model predictions from QVQ and identified four distinct error types that hinder productive reflection. These patterns are illustrated in Fig.~\ref{fig:ref_error_example} and their distribution is shown in Fig.~\ref{fig:ref_error_distribution}.

The four major error types are:

\begin{itemize}
    \item \textbf{Ineffective Reflection.} The model arrives at an incorrect conclusion and, upon reflecting, continues to make incorrect adjustments. This is the most common error type and is also witnessed most frequently.
    \item \textbf{Incompleteness.} The model proposes new analytical approaches but does not execute them, only stopping at the initial thought. The reflection slows down the inference process without bringing any gain.
    \item \textbf{Repetition.} The model restates previous content or methods without introducing new insights, leading to inefficient reasoning.
    \item \textbf{Interference.} The model initially reaches a correct conclusion but, through reflection, introduces errors.
\end{itemize}

Understanding and mitigating these errors is crucial for improving the reliability of LMM reflection mechanisms. The analysis provides the opportunity to focus on solving specific error types to enhance the overall reflection quality.

\section{Conclusion}
In this paper, we have introduced MME-CoT, a comprehensive benchmark designed to evaluate Chain-of-Thought reasoning in Large Multimodal Models. 
Our dataset comprises six categories to cover most scenarios of visual reasoning tasks.
To gain a thorough understanding of the reasoning process, we design a novel CoT evaluation suite with three metrics. 
Our systematic evaluation obtains useful insights into the issues within the current state-of-the-art Large Multimodal Models.
We identify critical flaws in all the tested open-source models.
As the field continues to evolve, MME-CoT stands as a valuable tool for measuring progress and identifying areas for improvement in the development of more sophisticated multimodal AI systems.

\section*{Impact Statement}
This paper presents work whose goal is to advance the field
of Computer Vision and Machine Learning. There are many
potential societal consequences of our work, none of which
we feel must be specifically highlighted here.

\bibliography{example_paper}

\begin{thebibliography}{57}
\providecommand{\natexlab}[1]{#1}
\providecommand{\url}[1]{\texttt{#1}}
\expandafter\ifx\csname urlstyle\endcsname\relax
  \providecommand{\doi}[1]{doi: #1}\else
  \providecommand{\doi}{doi: \begingroup \urlstyle{rm}\Url}\fi

\bibitem[Chen et~al.(2023)Chen, Zheng, Wang, Xu, Huang, Pan, Wang, Wang, Qiao, Lu, et~al.]{chen2023videollm}
Chen, G., Zheng, Y.-D., Wang, J., Xu, J., Huang, Y., Pan, J., Wang, Y., Wang, Y., Qiao, Y., Lu, T., et~al.
\newblock Videollm: Modeling video sequence with large language models.
\newblock \emph{arXiv preprint arXiv:2305.13292}, 2023.

\bibitem[Chen et~al.(2024{\natexlab{a}})Chen, Qin, Zhang, Chen, Xu, and Che]{chen-etal-2024-m3cot}
Chen, Q., Qin, L., Zhang, J., Chen, Z., Xu, X., and Che, W.
\newblock M$^3$cot: A novel benchmark for multi-domain multi-step multi-modal chain-of-thought.
\newblock In \emph{Proc. of ACL}, 2024{\natexlab{a}}.

\bibitem[Chen et~al.(2024{\natexlab{b}})Chen, Wang, Cao, Liu, Gao, Cui, Zhu, Ye, Tian, Liu, et~al.]{chen2024expanding}
Chen, Z., Wang, W., Cao, Y., Liu, Y., Gao, Z., Cui, E., Zhu, J., Ye, S., Tian, H., Liu, Z., et~al.
\newblock Expanding performance boundaries of open-source multimodal models with model, data, and test-time scaling.
\newblock \emph{arXiv preprint arXiv:2412.05271}, 2024{\natexlab{b}}.

\bibitem[Chen et~al.(2024{\natexlab{c}})Chen, Wang, Tian, Ye, Gao, Cui, Tong, Hu, Luo, Ma, et~al.]{chen2024far}
Chen, Z., Wang, W., Tian, H., Ye, S., Gao, Z., Cui, E., Tong, W., Hu, K., Luo, J., Ma, Z., et~al.
\newblock How far are we to gpt-4v? closing the gap to commercial multimodal models with open-source suites.
\newblock \emph{arXiv preprint arXiv:2404.16821}, 2024{\natexlab{c}}.

\bibitem[Du et~al.(2025)Du, Liu, Li, Zhao, Huo, Wang, Chen, Liu, Wang, and Wen]{du2025virgo}
Du, Y., Liu, Z., Li, Y., Zhao, W.~X., Huo, Y., Wang, B., Chen, W., Liu, Z., Wang, Z., and Wen, J.-R.
\newblock Virgo: A preliminary exploration on reproducing o1-like mllm.
\newblock \emph{arXiv preprint arXiv:2501.01904}, 2025.

\bibitem[Duan et~al.(2024)Duan, Yang, Qiao, Fang, Chen, Liu, Dong, Zang, Zhang, Wang, et~al.]{duan2024vlmevalkit}
Duan, H., Yang, J., Qiao, Y., Fang, X., Chen, L., Liu, Y., Dong, X., Zang, Y., Zhang, P., Wang, J., et~al.
\newblock Vlmevalkit: An open-source toolkit for evaluating large multi-modality models.
\newblock In \emph{Proceedings of the 32nd ACM International Conference on Multimedia}, pp.\  11198--11201, 2024.

\bibitem[Gao et~al.(2024)Gao, Zhang, Liu, Qiu, Huang, Lin, Zhao, Geng, Lin, Jin, et~al.]{gao2024sphinx}
Gao, P., Zhang, R., Liu, C., Qiu, L., Huang, S., Lin, W., Zhao, S., Geng, S., Lin, Z., Jin, P., et~al.
\newblock Sphinx-x: Scaling data and parameters for a family of multi-modal large language models.
\newblock \emph{ICML 2024}, 2024.

\bibitem[Golovneva et~al.(2022)Golovneva, Chen, Poff, Corredor, Zettlemoyer, Fazel-Zarandi, and Celikyilmaz]{golovneva2022roscoe}
Golovneva, O., Chen, M., Poff, S., Corredor, M., Zettlemoyer, L., Fazel-Zarandi, M., and Celikyilmaz, A.
\newblock Roscoe: A suite of metrics for scoring step-by-step reasoning.
\newblock \emph{arXiv preprint arXiv:2212.07919}, 2022.

\bibitem[Guo et~al.(2025{\natexlab{a}})Guo, Yang, Zhang, Song, Zhang, Xu, Zhu, Ma, Wang, Bi, et~al.]{guo2025deepseek}
Guo, D., Yang, D., Zhang, H., Song, J., Zhang, R., Xu, R., Zhu, Q., Ma, S., Wang, P., Bi, X., et~al.
\newblock Deepseek-r1: Incentivizing reasoning capability in llms via reinforcement learning.
\newblock \emph{arXiv preprint arXiv:2501.12948}, 2025{\natexlab{a}}.

\bibitem[Guo et~al.(2023)Guo, Zhang, Zhu, Tang, Ma, Han, Chen, Gao, Li, Li, et~al.]{guo2023point}
Guo, Z., Zhang, R., Zhu, X., Tang, Y., Ma, X., Han, J., Chen, K., Gao, P., Li, X., Li, H., et~al.
\newblock Point-bind \& point-llm: Aligning point cloud with multi-modality for 3d understanding, generation, and instruction following.
\newblock \emph{arXiv preprint arXiv:2309.00615}, 2023.

\bibitem[Guo et~al.(2024{\natexlab{a}})Guo, Zhang, Chen, Gao, Gao, Li, and Heng]{sciverse}
Guo, Z., Zhang, R., Chen, H., Gao, J., Gao, P., Li, H., and Heng, P.-A.
\newblock Sciverse.
\newblock \emph{https://sciverse-cuhk.github.io}, 2024{\natexlab{a}}.
\newblock URL \url{https://sciverse-cuhk.github.io/}.

\bibitem[Guo et~al.(2024{\natexlab{b}})Guo, Zhang, Zhu, Tong, Gao, Li, and Heng]{guo2024sam2point}
Guo, Z., Zhang, R., Zhu, X., Tong, C., Gao, P., Li, C., and Heng, P.-A.
\newblock Sam2point: Segment any 3d as videos in zero-shot and promptable manners.
\newblock \emph{arXiv preprint arXiv:2408.16768}, 2024{\natexlab{b}}.

\bibitem[Guo et~al.(2025{\natexlab{b}})Guo, Zhang, Tong, Zhao, Gao, Li, and Heng]{guo2025can}
Guo, Z., Zhang, R., Tong, C., Zhao, Z., Gao, P., Li, H., and Heng, P.-A.
\newblock Can we generate images with cot? let's verify and reinforce image generation step by step.
\newblock \emph{arXiv preprint arXiv:2501.13926}, 2025{\natexlab{b}}.

\bibitem[Hao et~al.(2024)Hao, Gu, Luo, Liu, Shao, Wang, Xie, Ma, Samavedhi, Gao, et~al.]{hao2024llm}
Hao, S., Gu, Y., Luo, H., Liu, T., Shao, X., Wang, X., Xie, S., Ma, H., Samavedhi, A., Gao, Q., et~al.
\newblock Llm reasoners: New evaluation, library, and analysis of step-by-step reasoning with large language models.
\newblock \emph{arXiv preprint arXiv:2404.05221}, 2024.

\bibitem[He et~al.(2024)He, Luo, Bai, Hu, Thai, Shen, Hu, Han, Huang, Zhang, Liu, Qi, Liu, and Sun]{he2024olympiadbench}
He, C., Luo, R., Bai, Y., Hu, S., Thai, Z.~L., Shen, J., Hu, J., Han, X., Huang, Y., Zhang, Y., Liu, J., Qi, L., Liu, Z., and Sun, M.
\newblock Olympiadbench: A challenging benchmark for promoting agi with olympiad-level bilingual multimodal scientific problems, 2024.

\bibitem[Jia et~al.(2024)Jia, Liu, Chen, Gu, Wang, Luo, Lee, Wang, Wang, Zhang, et~al.]{jia2024lift3d}
Jia, Y., Liu, J., Chen, S., Gu, C., Wang, Z., Luo, L., Lee, L., Wang, P., Wang, Z., Zhang, R., et~al.
\newblock Lift3d foundation policy: Lifting 2d large-scale pretrained models for robust 3d robotic manipulation.
\newblock \emph{arXiv preprint arXiv:2411.18623}, 2024.

\bibitem[Jiang et~al.(2024)Jiang, Zhang, Guo, Wu, Lei, Qiu, Lu, Chen, Song, Gao, et~al.]{jiang2024mmsearch}
Jiang, D., Zhang, R., Guo, Z., Wu, Y., Lei, J., Qiu, P., Lu, P., Chen, Z., Song, G., Gao, P., et~al.
\newblock Mmsearch: Benchmarking the potential of large models as multi-modal search engines.
\newblock \emph{arXiv preprint arXiv:2409.12959}, 2024.

\bibitem[Li et~al.(2024{\natexlab{a}})Li, Zhang, Guo, Zhang, Li, Zhang, Zhang, Li, Liu, and Li]{li2024llava-ov}
Li, B., Zhang, Y., Guo, D., Zhang, R., Li, F., Zhang, H., Zhang, K., Li, Y., Liu, Z., and Li, C.
\newblock Llava-onevision: Easy visual task transfer.
\newblock \emph{arXiv preprint arXiv:2408.03326}, 2024{\natexlab{a}}.

\bibitem[Li et~al.(2024{\natexlab{b}})Li, Zhang, Zhang, Zhang, Li, Li, Ma, and Li]{li2024llava}
Li, F., Zhang, R., Zhang, H., Zhang, Y., Li, B., Li, W., Ma, Z., and Li, C.
\newblock Llava-next-interleave: Tackling multi-image, video, and 3d in large multimodal models.
\newblock \emph{arXiv preprint arXiv:2407.07895}, 2024{\natexlab{b}}.

\bibitem[Li et~al.(2022)Li, Li, Xiong, and Hoi]{li2022blip}
Li, J., Li, D., Xiong, C., and Hoi, S.
\newblock Blip: Bootstrapping language-image pre-training for unified vision-language understanding and generation.
\newblock In \emph{International Conference on Machine Learning}, pp.\  12888--12900. PMLR, 2022.

\bibitem[Li et~al.(2023)Li, He, Wang, Li, Wang, Luo, Wang, Wang, and Qiao]{li2023videochat}
Li, K., He, Y., Wang, Y., Li, Y., Wang, W., Luo, P., Wang, Y., Wang, L., and Qiao, Y.
\newblock Videochat: Chat-centric video understanding.
\newblock \emph{arXiv preprint arXiv:2305.06355}, 2023.

\bibitem[Lin et~al.(2023)Lin, Liu, Zhang, Gao, Qiu, Xiao, Qiu, Lin, Shao, Chen, et~al.]{lin2023sphinx}
Lin, Z., Liu, C., Zhang, R., Gao, P., Qiu, L., Xiao, H., Qiu, H., Lin, C., Shao, W., Chen, K., et~al.
\newblock Sphinx: The joint mixing of weights, tasks, and visual embeddings for multi-modal large language models.
\newblock \emph{ECCV 2024}, 2023.

\bibitem[Liu et~al.(2023)Liu, Li, Wu, and Lee]{liu2023llava}
Liu, H., Li, C., Wu, Q., and Lee, Y.~J.
\newblock Visual instruction tuning.
\newblock In \emph{NeurIPS}, 2023.

\bibitem[Lu et~al.(2023)Lu, Bansal, Xia, Liu, yue Li, Hajishirzi, Cheng, Chang, Galley, and Gao]{Lu2023MathVistaEM}
Lu, P., Bansal, H., Xia, T., Liu, J., yue Li, C., Hajishirzi, H., Cheng, H., Chang, K.-W., Galley, M., and Gao, J.
\newblock Mathvista: Evaluating math reasoning in visual contexts with gpt-4v, bard, and other large multimodal models.
\newblock \emph{ArXiv}, abs/2310.02255, 2023.

\bibitem[OpenAI(2023)]{openai2023gpt4v}
OpenAI.
\newblock {GPT-4V(ision)} system card, 2023.
\newblock URL \url{https://openai.com/research/gpt-4v-system-card}.

\bibitem[{OpenAI}(2024)]{gpt4omini}
{OpenAI}.
\newblock Gpt-4o mini: advancing cost-efficient intelligence.
\newblock \url{https://openai.com/index/gpt-4o-mini-advancing-cost-efficient-intelligence/}, 2024.

\bibitem[OpenAI(2024{\natexlab{a}})]{o1}
OpenAI.
\newblock Introducing openai o1, 2024., 2024{\natexlab{a}}.
\newblock URL \url{https://openai.com/o1/}.

\bibitem[OpenAI(2024{\natexlab{b}})]{openai2024gpt4o}
OpenAI.
\newblock Hello gpt-4o.
\newblock \url{https://openai.com/index/hello-gpt-4o/}, 2024{\natexlab{b}}.

\bibitem[Peng et~al.(2024)Peng, Li, Zhou, Xia, Zhang, Bai, Mao, Wang, He, Zhou, et~al.]{peng2024chimera}
Peng, T., Li, M., Zhou, H., Xia, R., Zhang, R., Bai, L., Mao, S., Wang, B., He, C., Zhou, A., et~al.
\newblock Chimera: Improving generalist model with domain-specific experts.
\newblock \emph{arXiv preprint arXiv:2412.05983}, 2024.

\bibitem[Prasad et~al.(2023)Prasad, Saha, Zhou, and Bansal]{prasad2023receval}
Prasad, A., Saha, S., Zhou, X., and Bansal, M.
\newblock Receval: Evaluating reasoning chains via correctness and informativeness.
\newblock \emph{arXiv preprint arXiv:2304.10703}, 2023.

\bibitem[{Qwen Team}(2024)]{Qwen2-VL}
{Qwen Team}.
\newblock Qwen2-vl.
\newblock 2024.

\bibitem[Radford et~al.(2021)Radford, Kim, Hallacy, Ramesh, Goh, Agarwal, Sastry, Askell, Mishkin, Clark, Krueger, and Sutskever]{Radford2021LearningTV}
Radford, A., Kim, J.~W., Hallacy, C., Ramesh, A., Goh, G., Agarwal, S., Sastry, G., Askell, A., Mishkin, P., Clark, J., Krueger, G., and Sutskever, I.
\newblock Learning transferable visual models from natural language supervision.
\newblock In \emph{International Conference on Machine Learning}, 2021.
\newblock URL \url{https://api.semanticscholar.org/CorpusID:231591445}.

\bibitem[Sprague et~al.(2024)Sprague, Yin, Rodriguez, Jiang, Wadhwa, Singhal, Zhao, Ye, Mahowald, and Durrett]{sprague2024cot}
Sprague, Z., Yin, F., Rodriguez, J.~D., Jiang, D., Wadhwa, M., Singhal, P., Zhao, X., Ye, X., Mahowald, K., and Durrett, G.
\newblock To cot or not to cot? chain-of-thought helps mainly on math and symbolic reasoning.
\newblock \emph{arXiv preprint arXiv:2409.12183}, 2024.

\bibitem[Team et~al.(2025)Team, Du, Gao, Xing, Jiang, Chen, Li, Xiao, Du, Liao, et~al.]{team2025kimi}
Team, K., Du, A., Gao, B., Xing, B., Jiang, C., Chen, C., Li, C., Xiao, C., Du, C., Liao, C., et~al.
\newblock Kimi k1. 5: Scaling reinforcement learning with llms.
\newblock \emph{arXiv preprint arXiv:2501.12599}, 2025.

\bibitem[Team(2024)]{qvq-72b-preview}
Team, Q.
\newblock Qvq: To see the world with wisdom, December 2024.
\newblock URL \url{https://qwenlm.github.io/blog/qvq-72b-preview/}.

\bibitem[Touvron et~al.(2023)Touvron, Lavril, Izacard, Martinet, Lachaux, Lacroix, Rozi{\`e}re, Goyal, Hambro, Azhar, et~al.]{touvron2023llama}
Touvron, H., Lavril, T., Izacard, G., Martinet, X., Lachaux, M.-A., Lacroix, T., Rozi{\`e}re, B., Goyal, N., Hambro, E., Azhar, F., et~al.
\newblock Llama: Open and efficient foundation language models.
\newblock \emph{arXiv preprint arXiv:2302.13971}, 2023.

\bibitem[Wang et~al.(2024{\natexlab{a}})Wang, Fu, Huang, Li, Liu, Liu, Ma, Xu, Zhou, Zhang, et~al.]{wang2024muirbench}
Wang, F., Fu, X., Huang, J.~Y., Li, Z., Liu, Q., Liu, X., Ma, M.~D., Xu, N., Zhou, W., Zhang, K., et~al.
\newblock Muirbench: A comprehensive benchmark for robust multi-image understanding.
\newblock \emph{arXiv preprint arXiv:2406.09411}, 2024{\natexlab{a}}.

\bibitem[Wang et~al.(2024{\natexlab{b}})Wang, Bai, Tan, Wang, Fan, Bai, Chen, Liu, Wang, Ge, et~al.]{wang2024qwen2}
Wang, P., Bai, S., Tan, S., Wang, S., Fan, Z., Bai, J., Chen, K., Liu, X., Wang, J., Ge, W., et~al.
\newblock Qwen2-vl: Enhancing vision-language model's perception of the world at any resolution.
\newblock \emph{arXiv preprint arXiv:2409.12191}, 2024{\natexlab{b}}.

\bibitem[Wang et~al.(2024{\natexlab{c}})Wang, Chen, Wang, Cao, Liu, Gao, Zhu, Zhu, Lu, Qiao, and Dai]{wang2024mpo}
Wang, W., Chen, Z., Wang, W., Cao, Y., Liu, Y., Gao, Z., Zhu, J., Zhu, X., Lu, L., Qiao, Y., and Dai, J.
\newblock Enhancing the reasoning ability of multimodal large language models via mixed preference optimization.
\newblock \emph{arXiv preprint arXiv:2411.10442}, 2024{\natexlab{c}}.

\bibitem[Wang et~al.(2024{\natexlab{d}})Wang, Chen, Wang, Cao, Liu, Gao, Zhu, Zhu, Lu, Qiao, et~al.]{wang2024enhancing}
Wang, W., Chen, Z., Wang, W., Cao, Y., Liu, Y., Gao, Z., Zhu, J., Zhu, X., Lu, L., Qiao, Y., et~al.
\newblock Enhancing the reasoning ability of multimodal large language models via mixed preference optimization.
\newblock \emph{arXiv preprint arXiv:2411.10442}, 2024{\natexlab{d}}.

\bibitem[Wang et~al.(2024{\natexlab{e}})Wang, Xia, He, Chen, Liu, Zhu, Liang, Wu, Liu, Malladi, Chevalier, Arora, and Chen]{wang2024charxiv}
Wang, Z., Xia, M., He, L., Chen, H., Liu, Y., Zhu, R., Liang, K., Wu, X., Liu, H., Malladi, S., Chevalier, A., Arora, S., and Chen, D.
\newblock Charxiv: Charting gaps in realistic chart understanding in multimodal llms.
\newblock \emph{arXiv preprint arXiv:2406.18521}, 2024{\natexlab{e}}.

\bibitem[Wei et~al.(2022)Wei, Wang, Schuurmans, Bosma, Xia, Chi, Le, Zhou, et~al.]{wei2022chain}
Wei, J., Wang, X., Schuurmans, D., Bosma, M., Xia, F., Chi, E., Le, Q.~V., Zhou, D., et~al.
\newblock Chain-of-thought prompting elicits reasoning in large language models.
\newblock \emph{Advances in neural information processing systems}, 35:\penalty0 24824--24837, 2022.

\bibitem[Xu et~al.(2024)Xu, Jin, Li, Song, Sun, and Yuan]{xu2024llavacot}
Xu, G., Jin, P., Li, H., Song, Y., Sun, L., and Yuan, L.
\newblock Llava-cot: Let vision language models reason step-by-step, 2024.
\newblock URL \url{https://arxiv.org/abs/2411.10440}.

\bibitem[Xu et~al.(2023)Xu, Wang, Wang, Chen, Pang, and Lin]{xu2023pointllm}
Xu, R., Wang, X., Wang, T., Chen, Y., Pang, J., and Lin, D.
\newblock Pointllm: Empowering large language models to understand point clouds.
\newblock \emph{arXiv preprint arXiv:2308.16911}, 2023.

\bibitem[Yang et~al.(2024)Yang, Yang, Hui, Zheng, Yu, Zhou, Li, Li, Liu, Huang, Dong, Wei, Lin, Tang, Wang, Yang, Tu, Zhang, Ma, Xu, Zhou, Bai, He, Lin, Dang, Lu, Chen, Yang, Li, Xue, Ni, Zhang, Wang, Peng, Men, Gao, Lin, Wang, Bai, Tan, Zhu, Li, Liu, Ge, Deng, Zhou, Ren, Zhang, Wei, Ren, Fan, Yao, Zhang, Wan, Chu, Liu, Cui, Zhang, and Fan]{qwen2}
Yang, A., Yang, B., Hui, B., Zheng, B., Yu, B., Zhou, C., Li, C., Li, C., Liu, D., Huang, F., Dong, G., Wei, H., Lin, H., Tang, J., Wang, J., Yang, J., Tu, J., Zhang, J., Ma, J., Xu, J., Zhou, J., Bai, J., He, J., Lin, J., Dang, K., Lu, K., Chen, K., Yang, K., Li, M., Xue, M., Ni, N., Zhang, P., Wang, P., Peng, R., Men, R., Gao, R., Lin, R., Wang, S., Bai, S., Tan, S., Zhu, T., Li, T., Liu, T., Ge, W., Deng, X., Zhou, X., Ren, X., Zhang, X., Wei, X., Ren, X., Fan, Y., Yao, Y., Zhang, Y., Wan, Y., Chu, Y., Liu, Y., Cui, Z., Zhang, Z., and Fan, Z.
\newblock Qwen2 technical report.
\newblock \emph{arXiv preprint arXiv:2407.10671}, 2024.

\bibitem[Yao et~al.(2024{\natexlab{a}})Yao, Huang, Wu, Zhang, Wang, Liu, Wang, Song, Feng, Shen, et~al.]{yao2024mulberry}
Yao, H., Huang, J., Wu, W., Zhang, J., Wang, Y., Liu, S., Wang, Y., Song, Y., Feng, H., Shen, L., et~al.
\newblock Mulberry: Empowering mllm with o1-like reasoning and reflection via collective monte carlo tree search.
\newblock \emph{arXiv preprint arXiv:2412.18319}, 2024{\natexlab{a}}.

\bibitem[Yao et~al.(2024{\natexlab{b}})Yao, Yu, Zhang, Wang, Cui, Zhu, Cai, Li, Zhao, He, et~al.]{yao2024minicpm}
Yao, Y., Yu, T., Zhang, A., Wang, C., Cui, J., Zhu, H., Cai, T., Li, H., Zhao, W., He, Z., et~al.
\newblock Minicpm-v: A gpt-4v level mllm on your phone.
\newblock \emph{arXiv preprint arXiv:2408.01800}, 2024{\natexlab{b}}.

\bibitem[Ying et~al.(2024)Ying, Meng, Wang, Li, Lin, Yang, Zhang, Zhang, Lin, Liu, et~al.]{ying2024mmt}
Ying, K., Meng, F., Wang, J., Li, Z., Lin, H., Yang, Y., Zhang, H., Zhang, W., Lin, Y., Liu, S., et~al.
\newblock Mmt-bench: A comprehensive multimodal benchmark for evaluating large vision-language models towards multitask agi.
\newblock \emph{arXiv preprint arXiv:2404.16006}, 2024.

\bibitem[Yu et~al.(2023)Yu, Yang, Li, Wang, Lin, Liu, Wang, and Wang]{yu2023mm}
Yu, W., Yang, Z., Li, L., Wang, J., Lin, K., Liu, Z., Wang, X., and Wang, L.
\newblock Mm-vet: Evaluating large multimodal models for integrated capabilities.
\newblock \emph{arXiv preprint arXiv:2308.02490}, 2023.

\bibitem[Yue et~al.(2024)Yue, Zheng, Ni, Wang, Zhang, Tong, Sun, Yu, Zhang, Sun, Su, Chen, and Neubig]{yue2024mmmuprorobustmultidisciplinemultimodal}
Yue, X., Zheng, T., Ni, Y., Wang, Y., Zhang, K., Tong, S., Sun, Y., Yu, B., Zhang, G., Sun, H., Su, Y., Chen, W., and Neubig, G.
\newblock Mmmu-pro: A more robust multi-discipline multimodal understanding benchmark, 2024.
\newblock URL \url{https://arxiv.org/abs/2409.02813}.

\bibitem[Zhang et~al.(2023)Zhang, Li, Li, Ren, Zou, Liu, Huang, Gao, Zhang, Li, et~al.]{zhang2023llava}
Zhang, H., Li, H., Li, F., Ren, T., Zou, X., Liu, S., Huang, S., Gao, J., Zhang, L., Li, C., et~al.
\newblock Llava-grounding: Grounded visual chat with large multimodal models.
\newblock \emph{arXiv preprint arXiv:2312.02949}, 2023.

\bibitem[Zhang et~al.(2024{\natexlab{a}})Zhang, Han, Liu, Zhou, Lu, Qiao, Li, and Gao]{zhang2024llama}
Zhang, R., Han, J., Liu, C., Zhou, A., Lu, P., Qiao, Y., Li, H., and Gao, P.
\newblock Llama-adapter: Efficient fine-tuning of large language models with zero-initialized attention.
\newblock In \emph{ICLR 2024}, 2024{\natexlab{a}}.

\bibitem[Zhang et~al.(2024{\natexlab{b}})Zhang, Han, Zhou, Hu, Yan, Lu, Li, Gao, and Qiao]{zhang2024llamaadapter}
Zhang, R., Han, J., Zhou, A., Hu, X., Yan, S., Lu, P., Li, H., Gao, P., and Qiao, Y.
\newblock {LL}a{MA}-adapter: Efficient fine-tuning of large language models with zero-initialized attention.
\newblock In \emph{The Twelfth International Conference on Learning Representations}, 2024{\natexlab{b}}.
\newblock URL \url{https://openreview.net/forum?id=d4UiXAHN2W}.

\bibitem[Zhang et~al.(2024{\natexlab{c}})Zhang, Jiang, Zhang, Lin, Guo, Qiu, Zhou, Lu, Chang, Gao, et~al.]{zhang2024mathverse}
Zhang, R., Jiang, D., Zhang, Y., Lin, H., Guo, Z., Qiu, P., Zhou, A., Lu, P., Chang, K.-W., Gao, P., et~al.
\newblock Mathverse: Does your multi-modal llm truly see the diagrams in visual math problems?
\newblock \emph{ECCV 2024}, 2024{\natexlab{c}}.

\bibitem[Zhang et~al.(2024{\natexlab{d}})Zhang, Wei, Jiang, Zhang, Guo, Tong, Liu, Zhou, Wei, Zhang, et~al.]{zhang2024mavis}
Zhang, R., Wei, X., Jiang, D., Zhang, Y., Guo, Z., Tong, C., Liu, J., Zhou, A., Wei, B., Zhang, S., et~al.
\newblock Mavis: Mathematical visual instruction tuning.
\newblock \emph{arXiv preprint arXiv:2407.08739}, 2024{\natexlab{d}}.

\bibitem[Zhang et~al.(2024{\natexlab{e}})Zhang, Bai, Zhang, Gu, Zhai, Susskind, and Jaitly]{zhang2024far}
Zhang, Y., Bai, H., Zhang, R., Gu, J., Zhai, S., Susskind, J., and Jaitly, N.
\newblock How far are we from intelligent visual deductive reasoning?
\newblock In \emph{COLM}, 2024{\natexlab{e}}.

\bibitem[Zhu et~al.(2023)Zhu, Chen, Shen, Li, and Elhoseiny]{zhu2023minigpt}
Zhu, D., Chen, J., Shen, X., Li, X., and Elhoseiny, M.
\newblock Minigpt-4: Enhancing vision-language understanding with advanced large language models.
\newblock \emph{arXiv preprint arXiv:2304.10592}, 2023.

\end{thebibliography}
\bibliographystyle{icml2025}

\newpage
\appendix
\onecolumn
\section*{Appendix Overview}
\begin{itemize}
    \item Section~\ref{appendix:related}: Related Work.
    \item Section~\ref{appendix:more_dataset}: More Dataset Details.
    \item Section~\ref{appendix:error_analysis}: Error Analysis.
    \item Section~\ref{appendix:more_qualitative}: More Qualitative Examples.
    \item Section~\ref{appendix:eval_setup}: Evaluation Prompts.
\end{itemize}

\section{Related Work}
\label{appendix:related}
\subsection{Large Multimodal Models}
The field of multimodal~\citep{Radford2021LearningTV, li2022blip, openai2023gpt4v, openai2024gpt4o} AI has experienced extraordinary growth, particularly through the development of Large Multimodal Models (LMMs)~\cite{liu2023llava,zhu2023minigpt,lin2023sphinx,Qwen2-VL}. These models build upon the achievements of Large Language Models (LLMs)~\citep{touvron2023llama,qwen2} and advanced vision models~\cite{Radford2021LearningTV}, expanding their capabilities to process multiple kinds of visual input~\cite{li2024llava,guo2023point,li2023videochat}.

Closed-source models, such as OpenAI's GPT-4o~\citep{openai2024gpt4o}, have demonstrated exceptional capabilities in visual understanding and reasoning. However, their closed-source nature creates barriers to widespread adoption and further development by the broader research community. In response, significant progress has been made in developing open-source alternatives. Early approaches like LLaVA~\cite{liu2023llava}, LLaMA-Adapter~\cite{zhang2024llamaadapter}, and MiniGPT-4~\cite{zhu2023minigpt} established a foundation by combining frozen CLIP models for image encoding with LLMs, enabling multimodal instruction tuning. Subsequent developments through projects such as InternVL2~\cite{chen2024far}, Qwen2-VL~\cite{Qwen2-VL}, SPHINX~\cite{gao2024sphinx,lin2023sphinx}, and MiniCPM-V~\cite{yao2024minicpm} have expanded these capabilities by incorporating more diverse visual instruction datasets and broadening application scenarios.

Recently, with the introduction of o1~\cite{o1}, the field of LMMs has also focused on enhancing the reasoning capability. \cite{wang2024enhancing} introduces mixed preference optimization with automatically constructed data. \cite{yao2024mulberry} proposes to leverage collective knowledge from multiple models to identify effective reasoning paths. Besides, several works~\cite{qvq-72b-preview,du2025virgo} have demonstrated the ability to replicate behaviors similar to o1 models, particularly regarding multi-step CoT reasoning with iterative self-reflection and verification processes.

\subsection{Reasoning Evaluation}
Several methods have been developed to evaluate reasoning in natural language processing, including ROSCOE~\cite{golovneva2022roscoe} and ReCEval~\cite{prasad2023receval}, which assess reasoning chains across multiple dimensions such as correctness and informativeness. However, these approaches are limited to text-only scenarios and do not address the unique challenges present in visual reasoning tasks. Furthermore, the emergence of long chain-of-thought (CoT) reasoning has introduced additional considerations, such as output efficiency and reflection quality, which existing evaluation methods do not adequately address.

On the other hand, various multimodal benchmarks have been developed to assess reasoning abilities across specific domains. Current exploration of visual reasoning predominantly focuses on the mathematics~\cite{zhang2024mavis,peng2024chimera} domains. 
MathVista~\cite{Lu2023MathVistaEM} provides a comprehensive collection of mathematical problems that assess mathematical and logical reasoning abilities. 
Building on this, MathVerse~\cite{zhang2024mathverse} introduces a new benchmark by eliminating redundant textual information to evaluate whether LMMs can accurately interpret graphical representations. 
OlympiadBench~\cite{he2024olympiadbench} further raises the complexity bar by incorporating challenging Olympiad-level mathematics and physics problems. Despite these advances in specialized domains, broader applications such as general-scene reasoning remain relatively unexplored.
Recent developments have begun to expand beyond purely scientific reasoning. For instance, M³CoT~\cite{chen-etal-2024-m3cot} and SciVerse~\cite{sciverse} incorporate commonsense tasks alongside scientific reasoning and knowledge-based assessment in the multimodal benchmark. However, most existing benchmarks focus solely on evaluating final answers while overlooking the intermediate steps, thus providing limited insights into the process through which models arrive at their conclusions.

\section{More Dataset Details}
\label{appendix:more_dataset}
\subsection{Data Source Distribution}
We visualize the data source distributions in our benchmark, which consists of 15 sets, including MathVerse~\cite{zhang2024mathverse}, MMMUPro~\cite{yue2024mmmuprorobustmultidisciplinemultimodal}, OlympiadBench~\cite{he2024olympiadbench}, MMT-Bench~\cite{ying2024mmt}, MuirBench~\cite{wang2024muirbench}, ml-rpm-bench~\cite{zhang2024far}, MMSearch~\cite{jiang2024mmsearch}, CharXiv~\cite{wang2024charxiv}, and SciVerse~\cite{sciverse}.

\begin{figure*}[!h]
\centering
\includegraphics[width=0.4\textwidth]{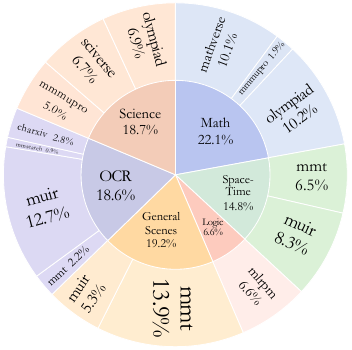} 
\caption{\textbf{Data Source Distribution of MME-CoT.}}
\label{appendix:more_dataset-source}
\end{figure*}

\newpage

\subsection{Preliminary Categorization Result}
\label{appendix:preliminary_result}
\begin{table}[htbp]
    \centering
    \caption{\textbf{Accuracy of MMT-Bench for different subcategories}. ACT: Action Understanding; AUT: Attribute Similarity; CNT: Cartoon Understanding; CIM: Counting; DOC: Diagram Understanding; EMO: Difference Spotting; HAL: Geographic Understanding; IIT: Image-Text Matching; IRT: Ordering; IQT: Scene Understanding; MEM: Visual Grounding; MIA: Visual Retrieval; OCR: Object Recognition; PLP: Physical Layout Prediction; RRE: Relationship Extraction; TMP: Temporal Reasoning; VCP: Visual Comprehension; VCR: Visual Coherence Reasoning; VGR: Visual Generation; VIL: Visual Identification; VPU: Visual Prediction Understanding; VRE: Visual Reasoning Evaluation.}
    \label{tab:hit_ratio}
    \setlength{\tabcolsep}{4pt} 
    \renewcommand{\arraystretch}{1.2}
    \small 
    \begin{tabularx}{\textwidth}{l *{22}{X}}
        \toprule
        File Name & 
        \rotatebox{90}{ACT} & \rotatebox{90}{AUT} & \rotatebox{90}{CNT} & \rotatebox{90}{CIM} & 
        \rotatebox{90}{DOC} & \rotatebox{90}{EMO} & \rotatebox{90}{HAL} & \rotatebox{90}{IIT} & 
        \rotatebox{90}{IRT} & \rotatebox{90}{IQT} & \rotatebox{90}{MEM} & \rotatebox{90}{MIA} & 
        \rotatebox{90}{OCR} & \rotatebox{90}{PLP} & \rotatebox{90}{RRE} & \rotatebox{90}{TMP} & 
        \rotatebox{90}{VCP} & \rotatebox{90}{VCR} & \rotatebox{90}{VGR} & \rotatebox{90}{VIL} & 
        \rotatebox{90}{VPU} & \rotatebox{90}{VRE} \\
        \midrule
        GPT4o-cot & 0.60 & 0.60 & 0.44 & 0.67 & 0.79 & 0.30 & 0.71 & 0.50 & 0.63 & 0.10 & 0.85 & 0.60 & 0.77 & 0.36 & 0.76 & 0.48 & 0.86 & 0.80 & 0.49 & 0.48 & 0.82 & 0.85 \\
        GPT4-direct & 0.53 & 0.60 & 0.44 & 0.67 & 0.81 & 0.23 & 0.69 & 0.33 & 0.66 & 0.25 & 0.80 & 0.43 & 0.78 & 0.42 & 0.78 & 0.36 & 0.89 & 0.85 & 0.41 & 0.37 & 0.85 & 0.85 \\
        Qwen2-VL-7B-cot & 0.53 & 0.61 & 0.34 & 0.65 & 0.77 & 0.53 & 0.74 & 0.40 & 0.31 & 0.20 & 0.78 & 0.58 & 0.60 & 0.43 & 0.69 & 0.43 & 0.85 & 0.90 & 0.54 & 0.35 & 0.79 & 0.81 \\
        Qwen2-VL-7B-direct & 0.49 & 0.67 & 0.40 & 0.78 & 0.75 & 0.52 & 0.73 & 0.43 & 0.31 & 0.10 & 0.78 & 0.55 & 0.60 & 0.54 & 0.69 & 0.40 & 0.85 & 0.85 & 0.67 & 0.38 & 0.85 & 0.82 \\
        \bottomrule
    \end{tabularx}
\end{table}

\begin{table}[htbp]
    \centering
    \caption{\textbf{Accuracy of MUIRBench for different subcategories}. AU: Action Understanding; AS: Attribute Similarity; CU: Cartoon Understanding; CO: Counting; DU: Diagram Understanding; DS: Difference Spotting; GU: Geographic Understanding; ITM: Image-Text Matching; OR: Ordering; SU: Scene Understanding; VG: Visual Grounding; VR: Visual Retrieval.}

    \label{tab:hit_ratio}
    \setlength{\tabcolsep}{4pt} 
    \renewcommand{\arraystretch}{1.2} 
    \small 
    \begin{tabularx}{\textwidth}{l XXXX XXXX XXXX XXXX}
        \toprule
        File Name & AU & AS & CU & CO & DU & DS & GU & ITM & OR & SU & VG & VR \\
        \midrule
        GPT4o-cot & 0.48 & 0.57 & 0.55 & 0.75 & 0.82 & 0.64 & 0.59 & 0.82 & 0.38 & 0.88 & 0.56 & 0.70 \\
        GPT4o-direct & 0.45 & 0.62 & 0.59 & 0.50 & 0.88 & 0.62 & 0.55 & 0.86 & 0.33 & 0.74 & 0.38 & 0.77 \\
        Qwen2-VL-7B-cot & 0.38 & 0.51 & 0.42 & 0.43 & 0.43 & 0.27 & 0.21 & 0.55 & 0.13 & 0.69 & 0.37 & 0.28 \\
        Qwen2-VL-7B-direct & 0.39 & 0.47 & 0.44 & 0.41 & 0.40 & 0.33 & 0.25 & 0.51 & 0.13 & 0.67 & 0.31 & 0.20 \\
        \bottomrule
    \end{tabularx}
\end{table}

\begin{table}[htbp]
    \centering
    \caption{\textbf{Accuracy of OlympiadBench for the mathematics and physics subcategories}.}
    \label{tab:hit_ratio_oe}
    \small 
    \begin{tabular}{lcc}
        \toprule
        File Name & Mathematics & Physics\\
        \midrule
        GPT4o-cot & 0.25 & 0.04 \\
        GPT4o-direct & 0.07 & 0.03 \\
        Qwen2-VL-7B-cot & 0.05 & 0.01 \\
        Qwen2-VL-7B-direct & 0.07 & 0.01 \\
        \bottomrule
    \end{tabular}
\end{table}

\newpage

\section{Error Analysis}
\label{appendix:error_analysis}
\begin{figure*}[!h]
\centering
\includegraphics[width=\textwidth]{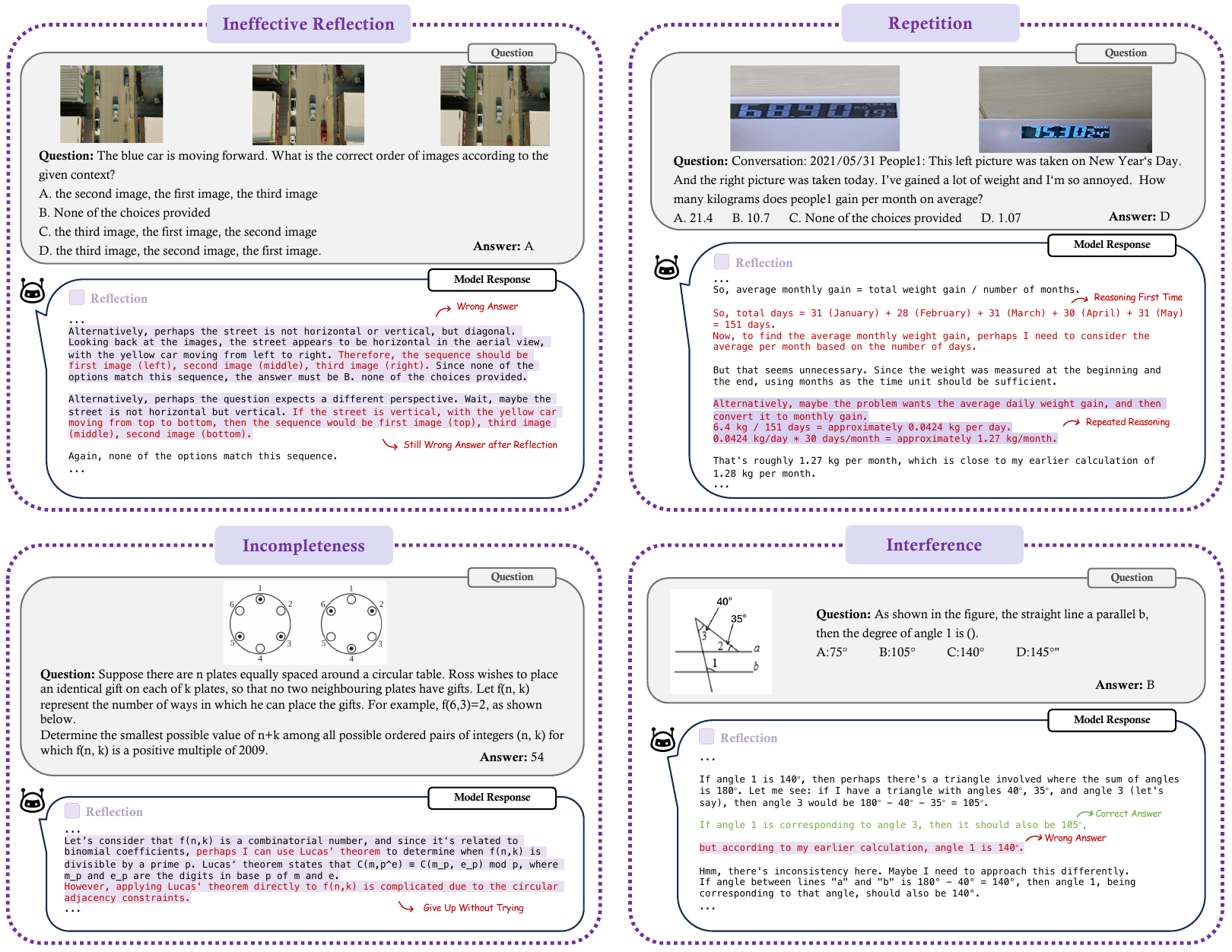} 
\caption{\textbf{Examples of Reflection Error Types.}}
\label{fig:ref_error_example}
\end{figure*}

We showcase the examples of the identified error types of reflection in Fig.~\ref{fig:ref_error_example}.

\newpage

\section{More Qualitative Examples}
\label{appendix:more_qualitative}
\begin{figure*}[!h]
\centering
\includegraphics[width=0.6\textwidth]{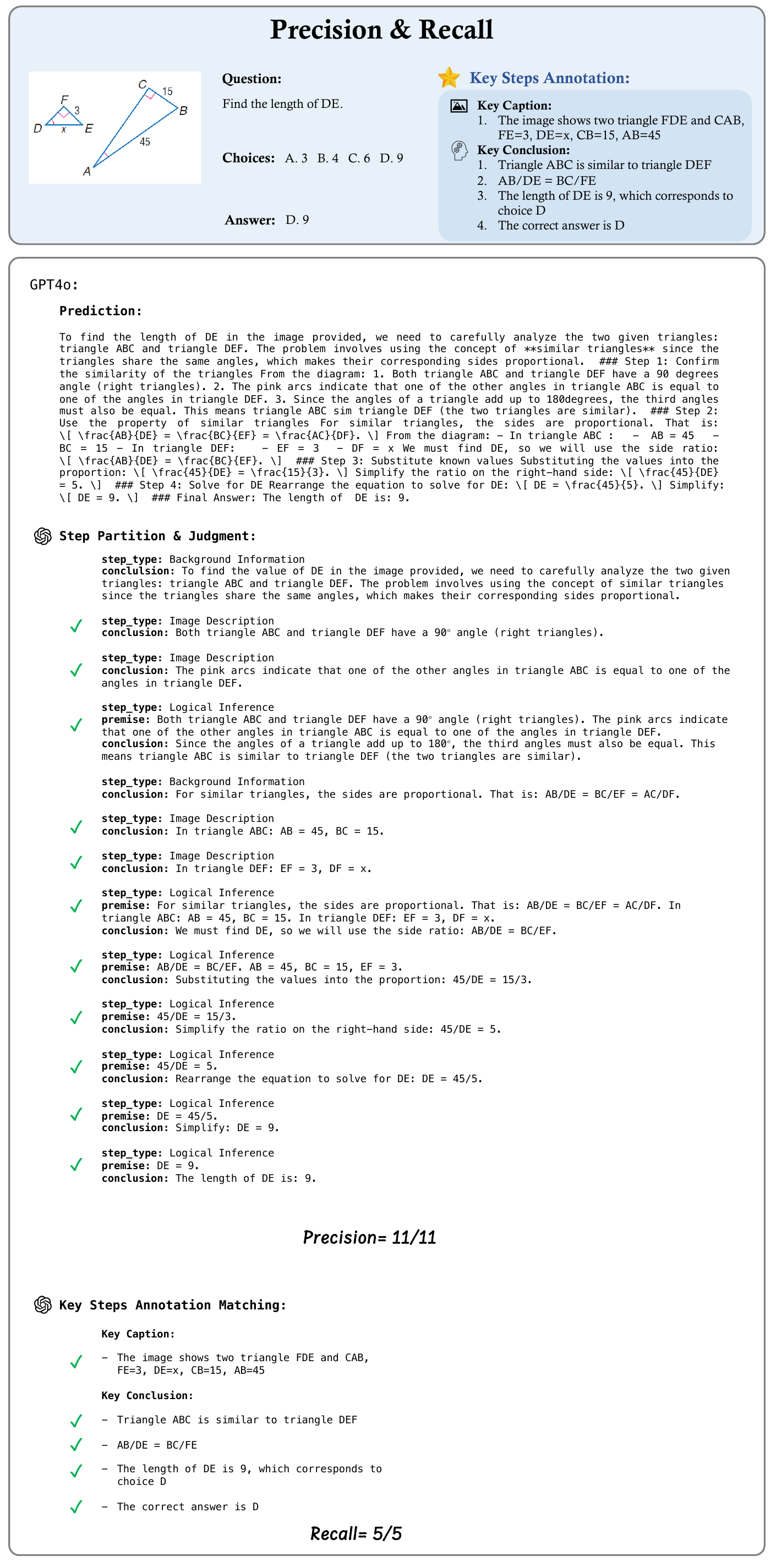} 
\caption{\textbf{Examples of Precision and Recall Evaluation.}}
\label{fig:precision_recall_example_GPT}
\end{figure*}
\newpage

\begin{figure*}[!h]
\centering
\includegraphics[width=0.9\textwidth]{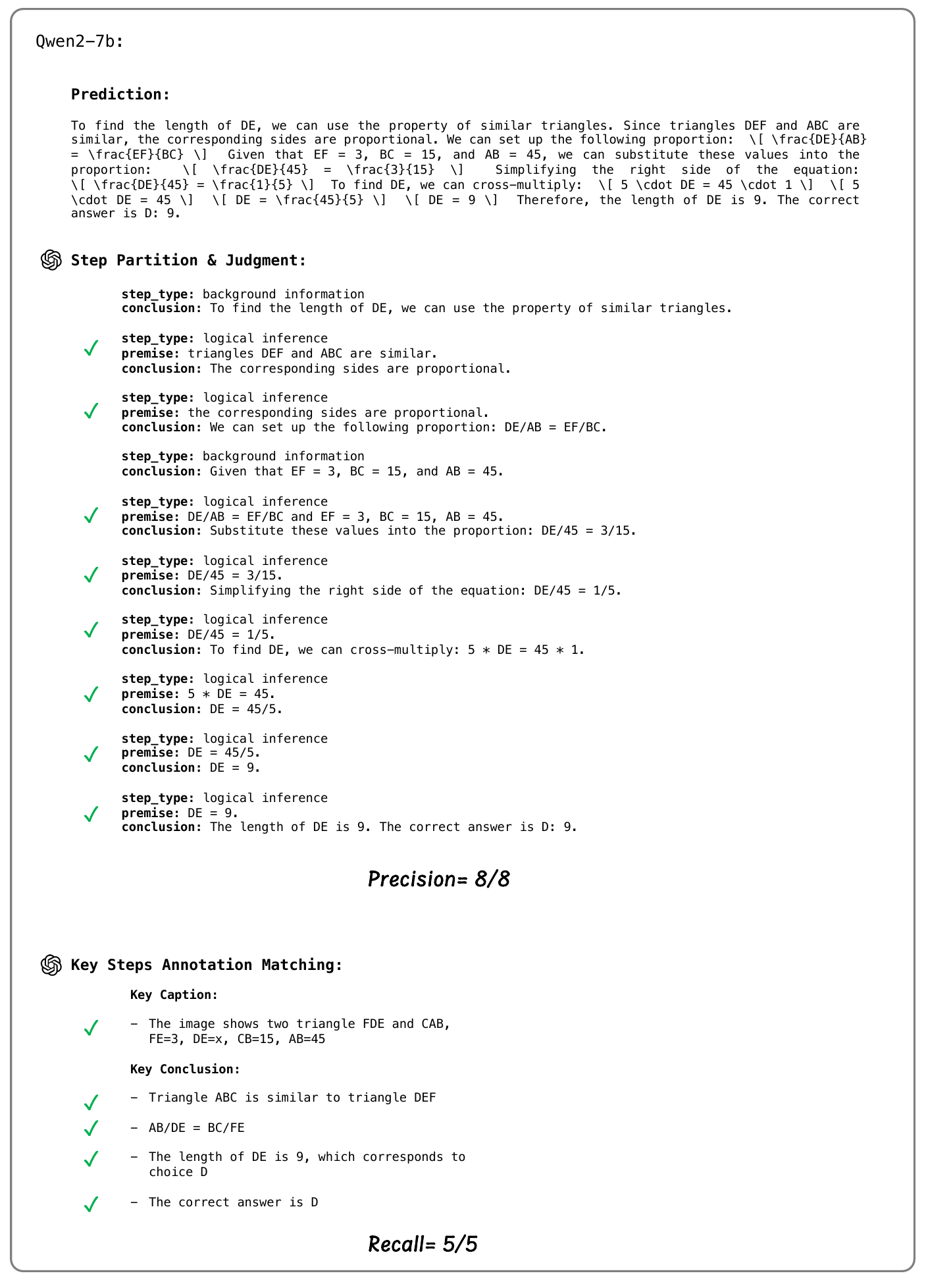} 
\caption{\textbf{Examples of Precision and Recall Evaluation.}}
\label{fig:precision_recall_example_Qwen}
\end{figure*}
\newpage

\begin{figure*}[!h]
\centering
\includegraphics[width=0.58\textwidth]{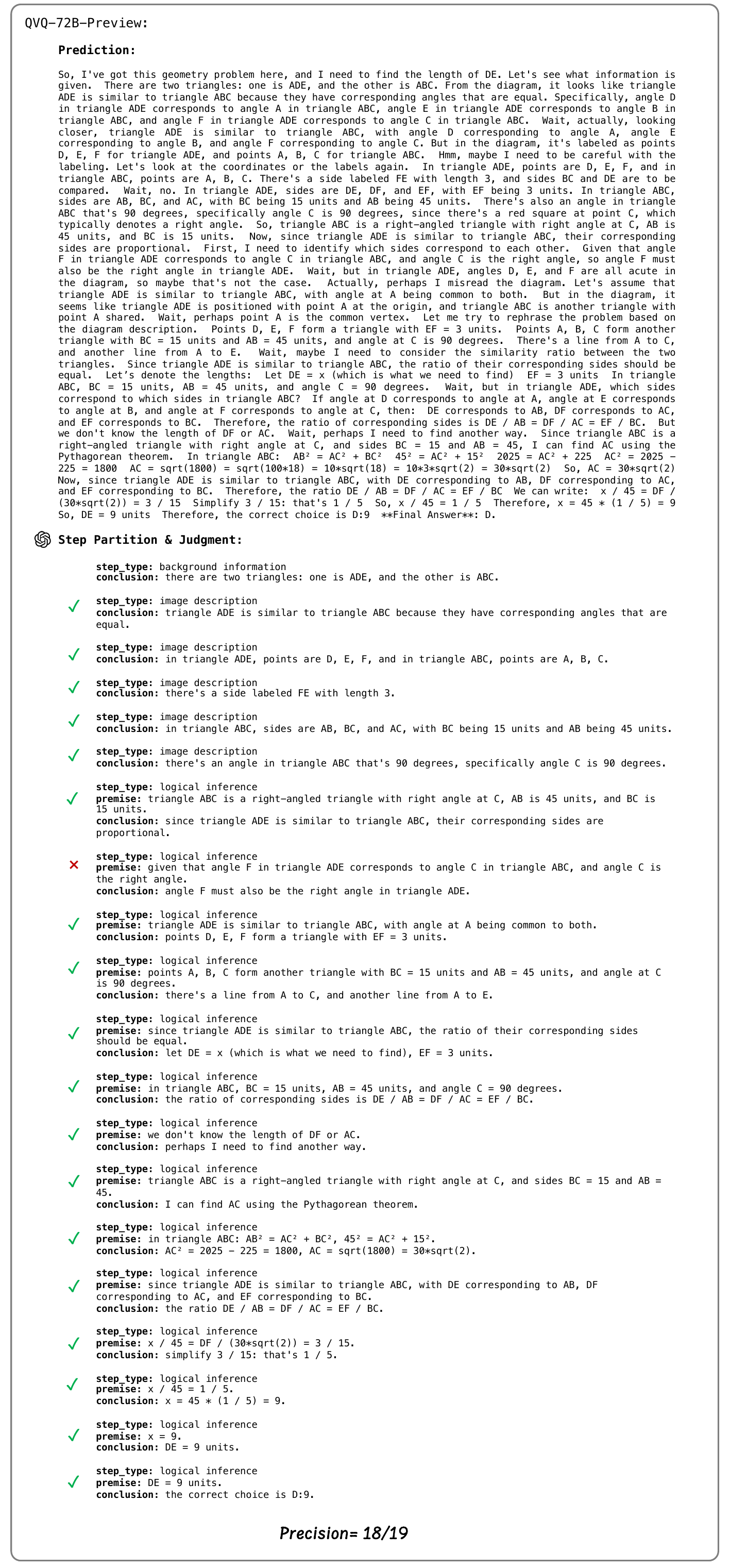}
\caption{\textbf{Examples of Precision and Recall Evaluation.}}
\label{fig:precision_recall_example_QVQ}
\end{figure*}
\newpage

\begin{figure*}[!h]
\centering
\includegraphics[width=\textwidth]{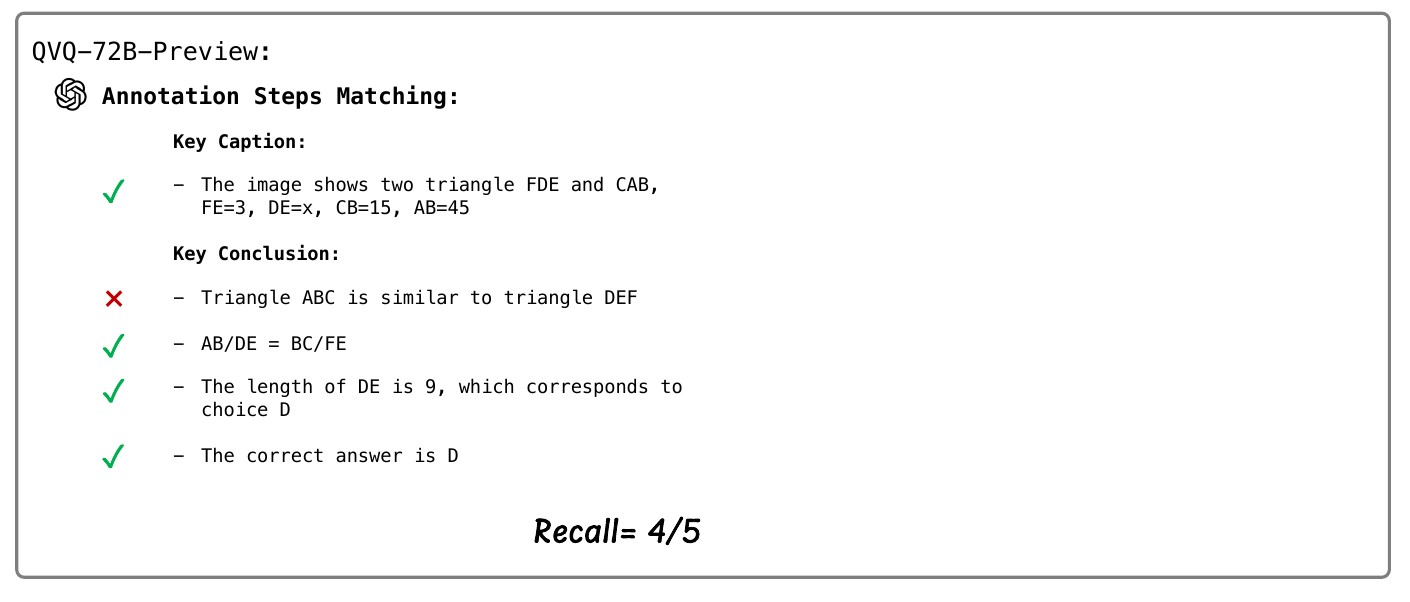} 
\caption{\textbf{Examples of Precision and Recall Evaluation.}}
\label{fig:precision_recall_example_QVQ2}
\end{figure*}
\newpage

\begin{figure*}[!h]
\centering
\includegraphics[width=0.51\textwidth]{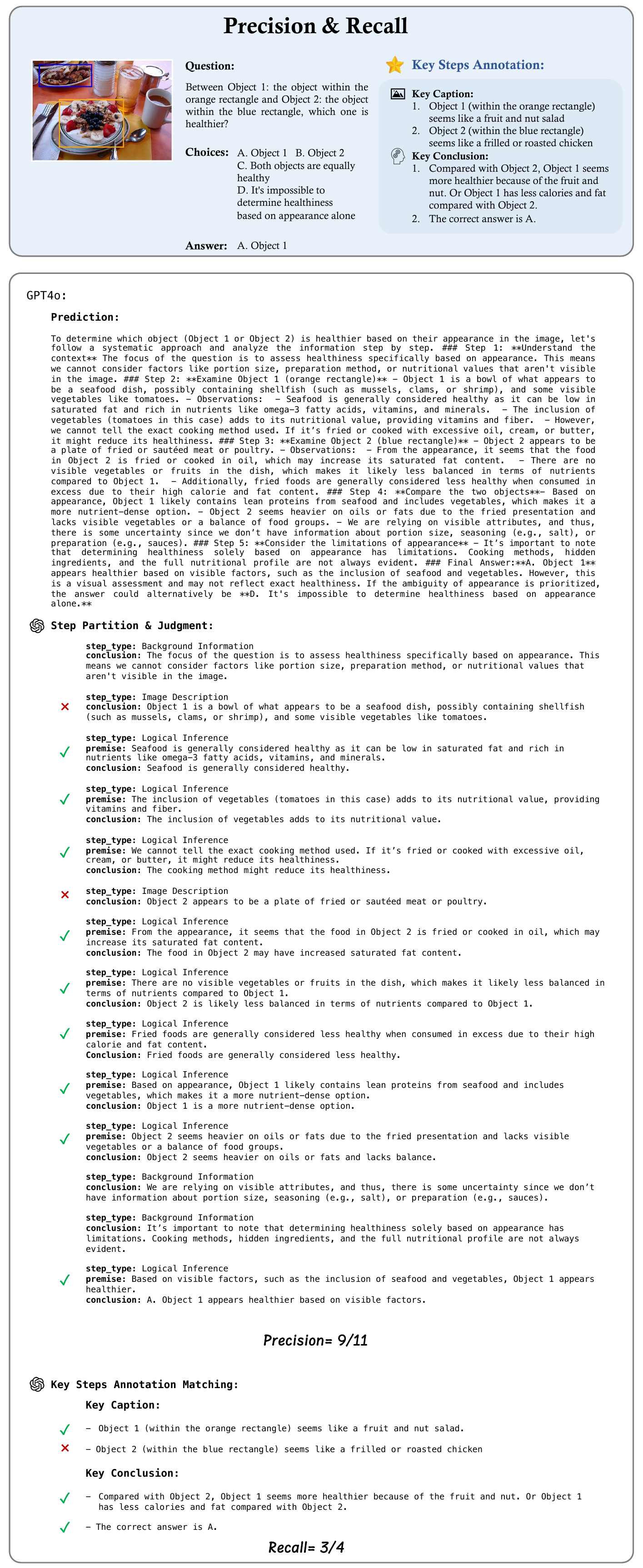} 
\caption{\textbf{Examples of Precision and Recall Evaluation.}}
\label{fig:precision_recall_example2_GPT}
\end{figure*}
\newpage

\begin{figure*}[!h]
\centering
\includegraphics[width=0.79\textwidth]{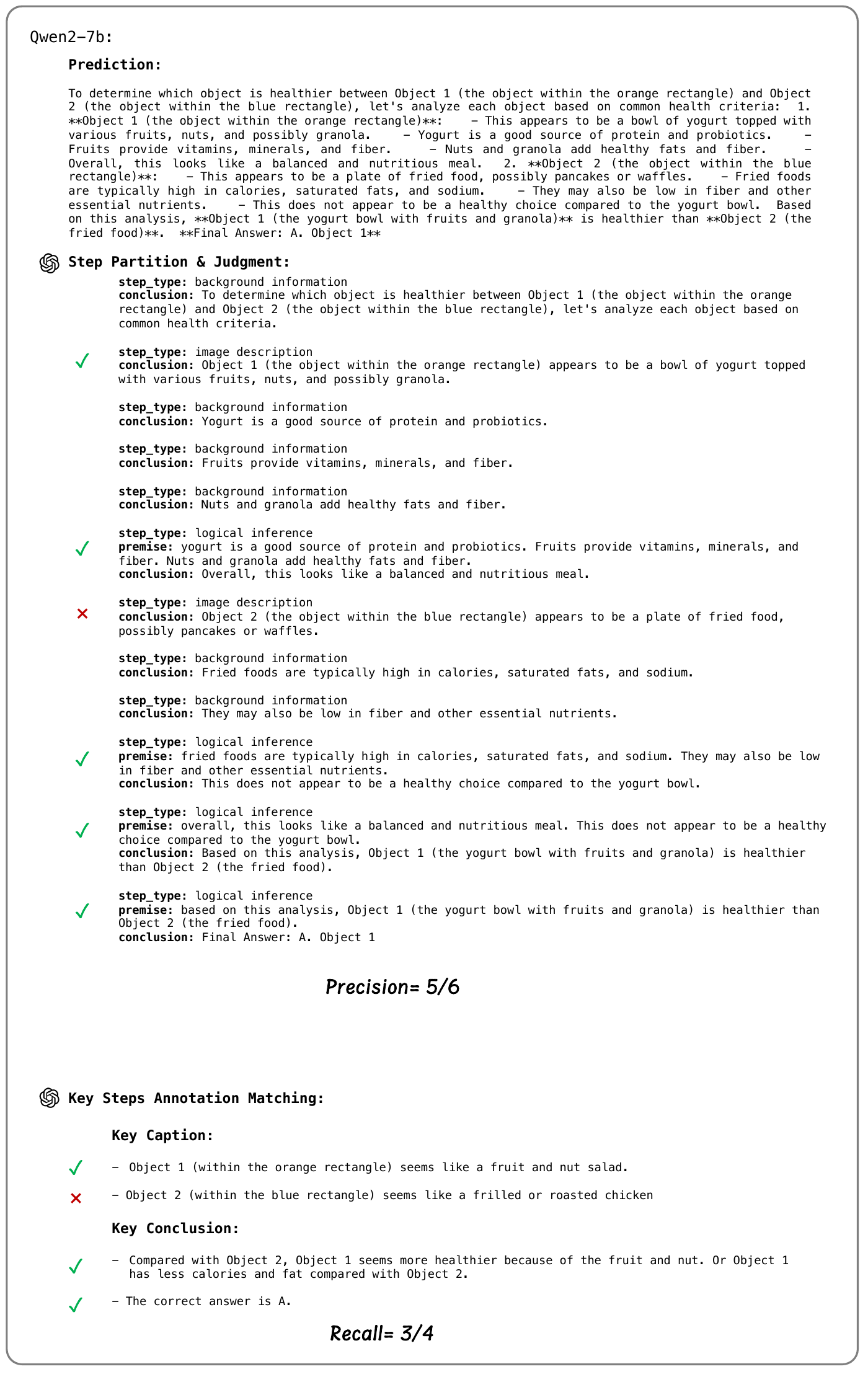} 
\caption{\textbf{Examples of Precision and Recall Evaluation.}}
\label{fig:precision_recall_example2_Qwen}
\end{figure*}
\newpage

\begin{figure*}[!h]
\centering
\includegraphics[width=0.81\textwidth]{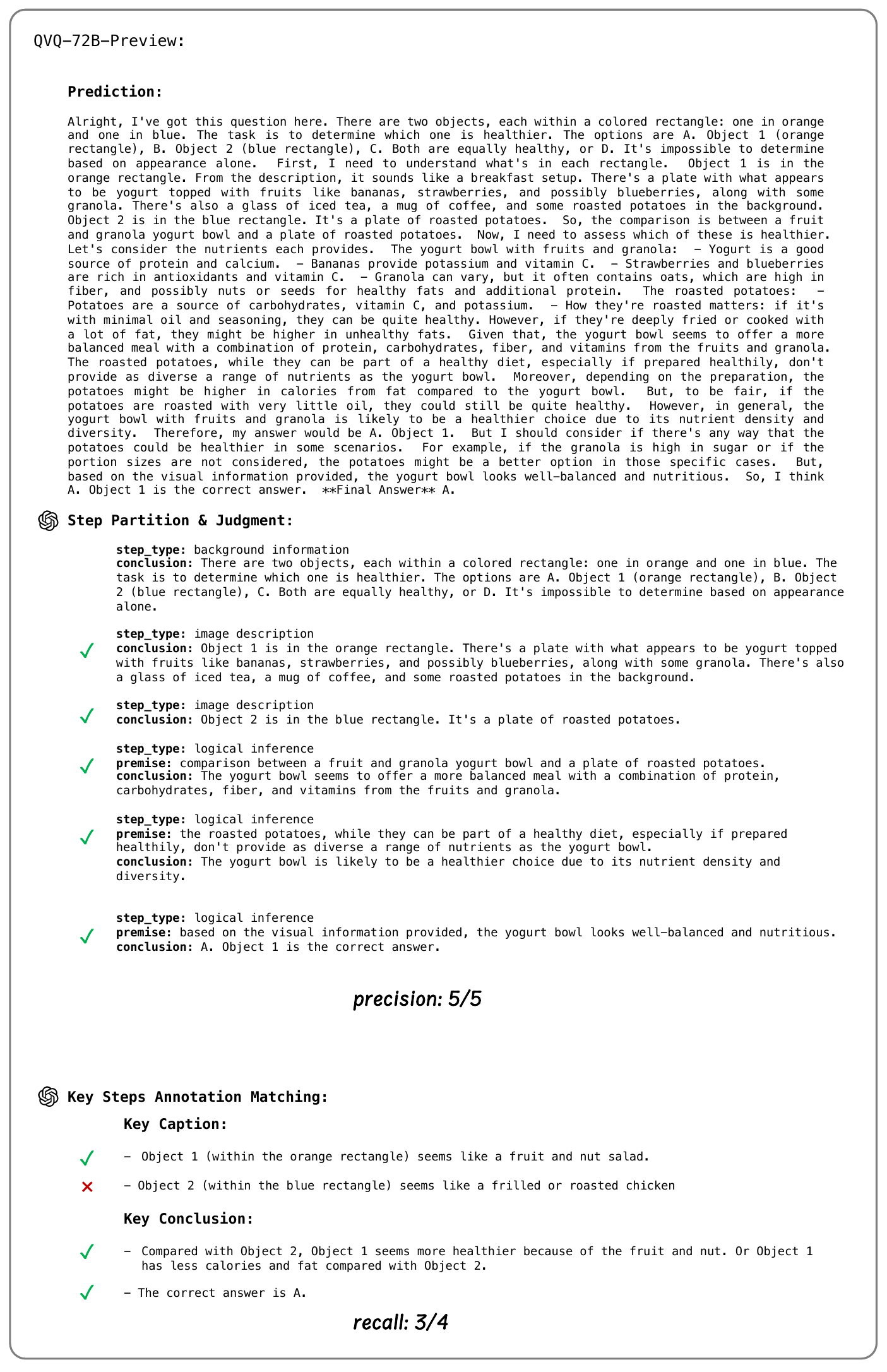} 
\caption{\textbf{Examples of Precision and Recall Evaluation.}}
\label{fig:precision_recall_example2_QVQ}
\end{figure*}
\newpage

\begin{figure*}[!h]
\centering
\includegraphics[width=\textwidth]{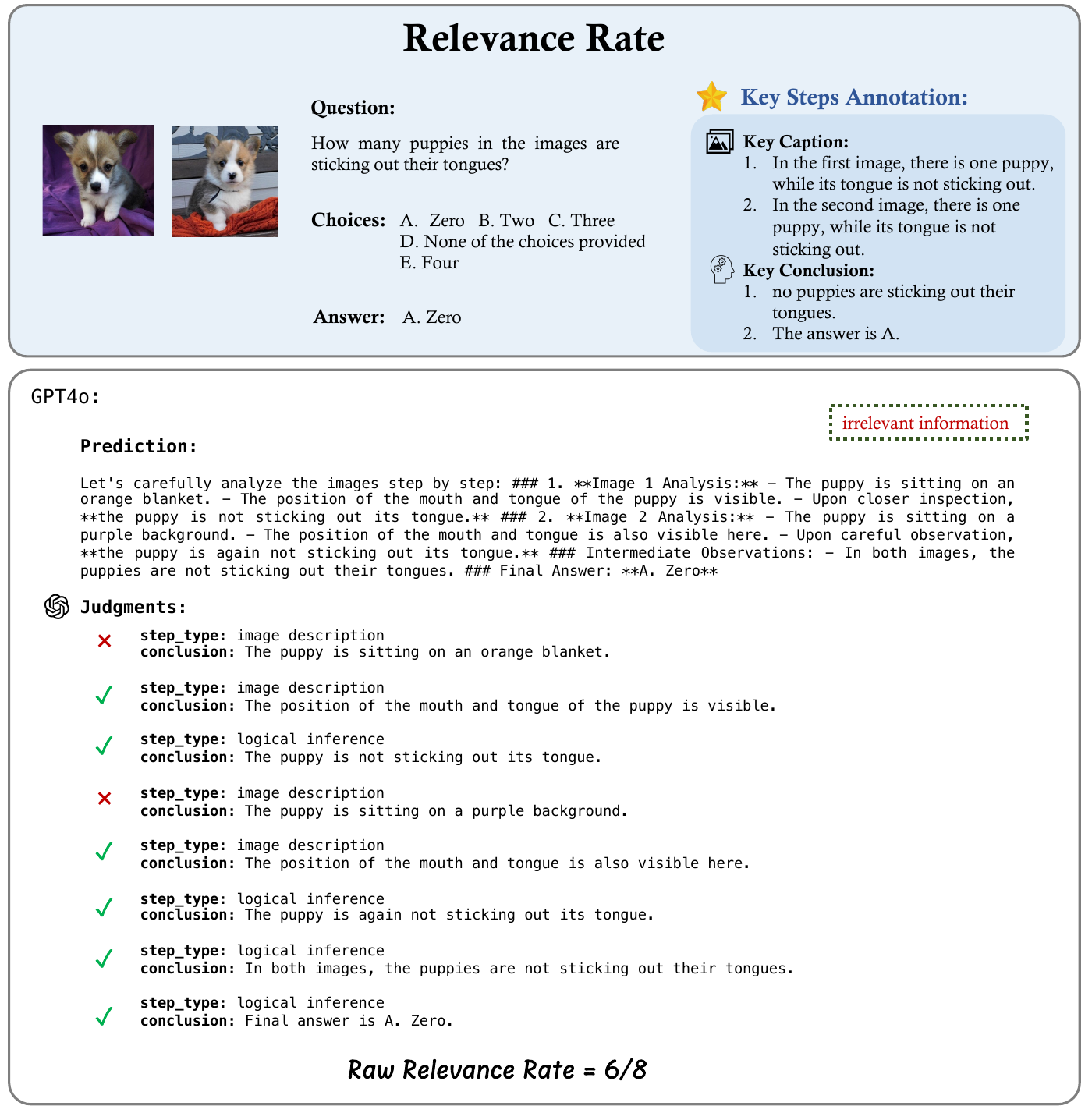} 
\caption{\textbf{Examples of Relevance Rate Evaluation.}}
\label{fig:relevance_example_GPT}
\end{figure*}
\newpage

\begin{figure*}[!h]
\centering
\includegraphics[width=\textwidth]{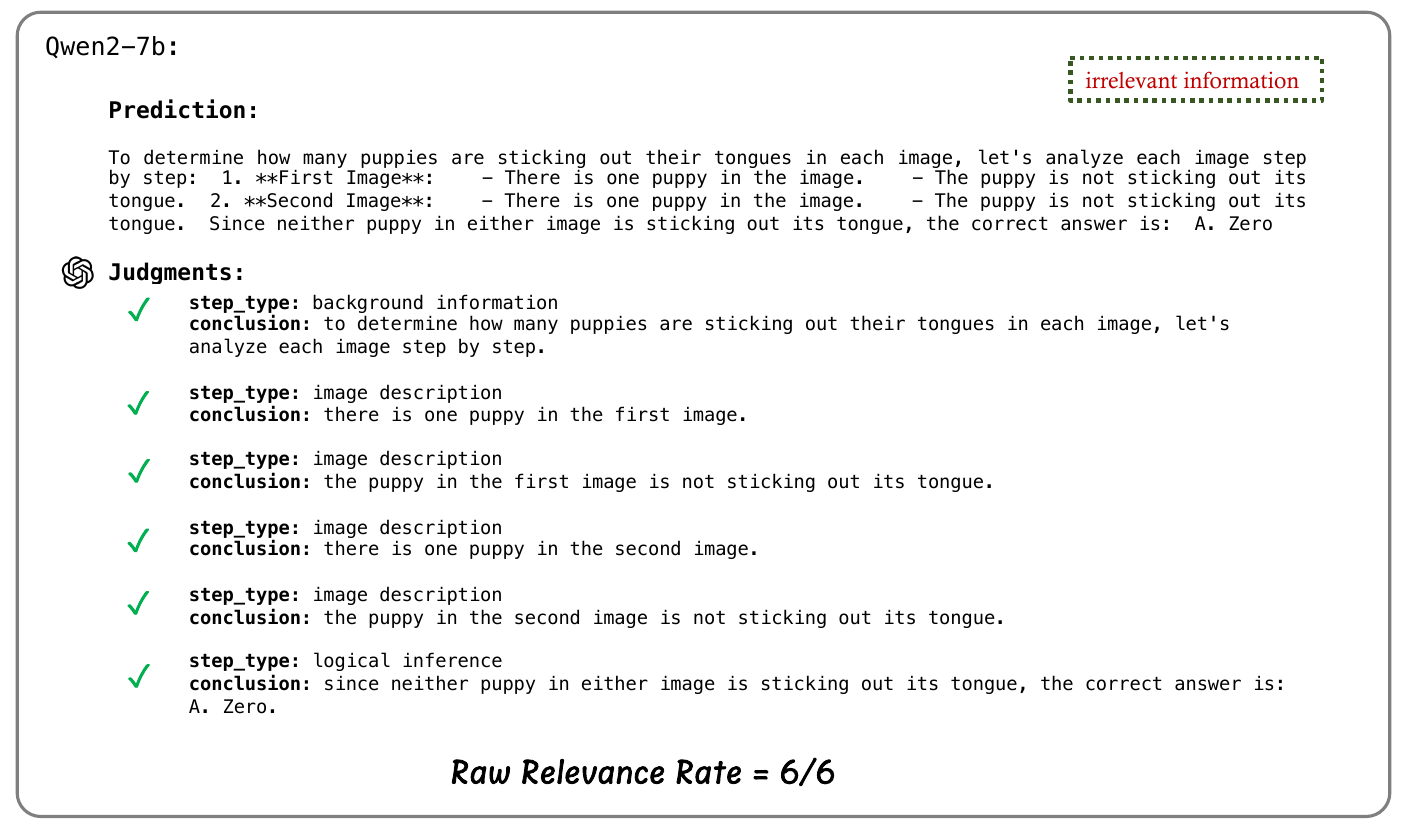} 
\caption{\textbf{Examples of Relevance Rate Evaluation.}}
\label{fig:relevance_example_Qwen}
\end{figure*}
\newpage

\begin{figure*}[!h]
\centering
\includegraphics[width=\textwidth]{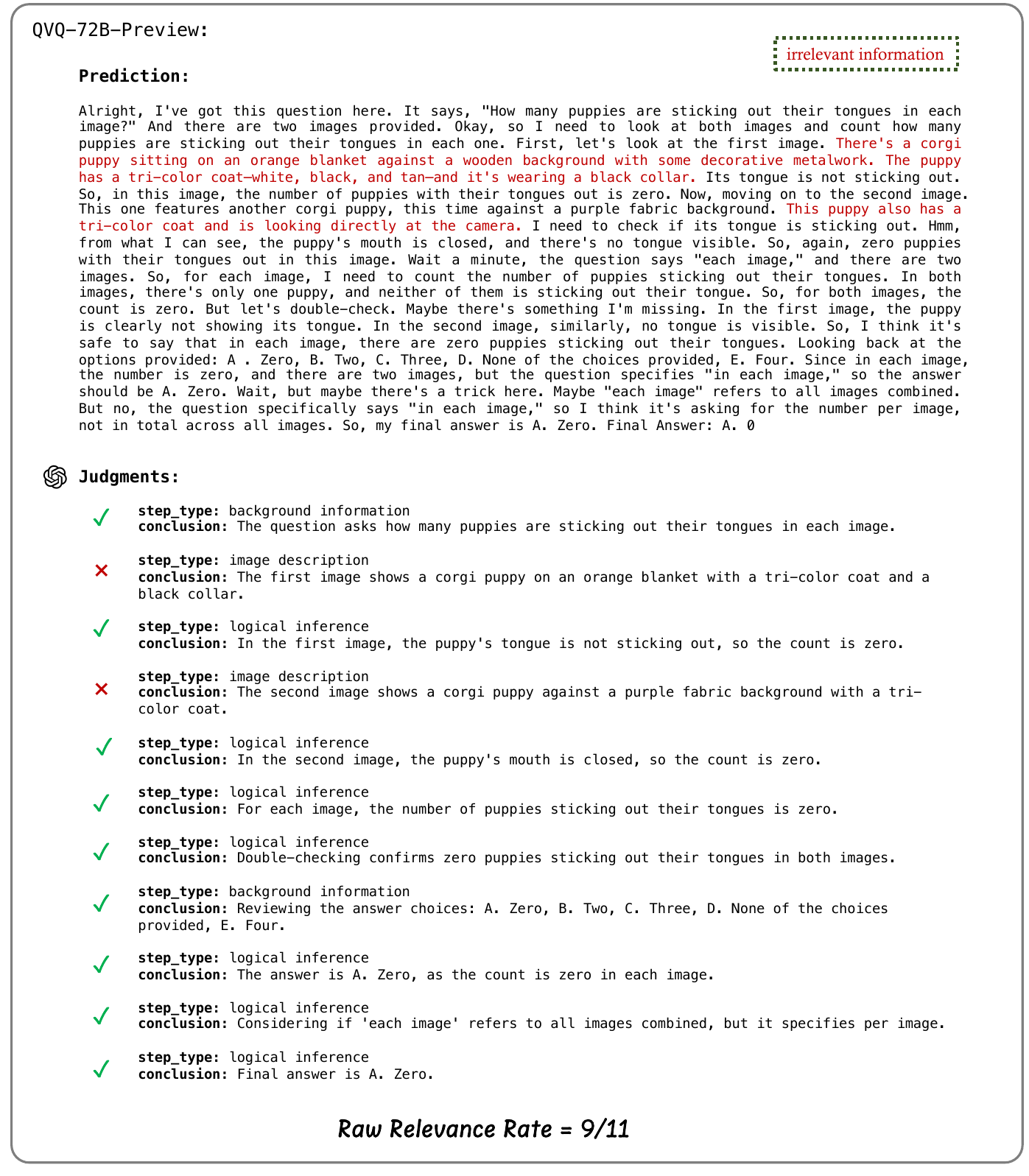} 
\caption{\textbf{Examples of Relevance Rate Evaluation.}}
\label{fig:relevance_example_QVQ}
\end{figure*}
\newpage

\begin{figure*}[!h]
\centering
\includegraphics[width=\textwidth]{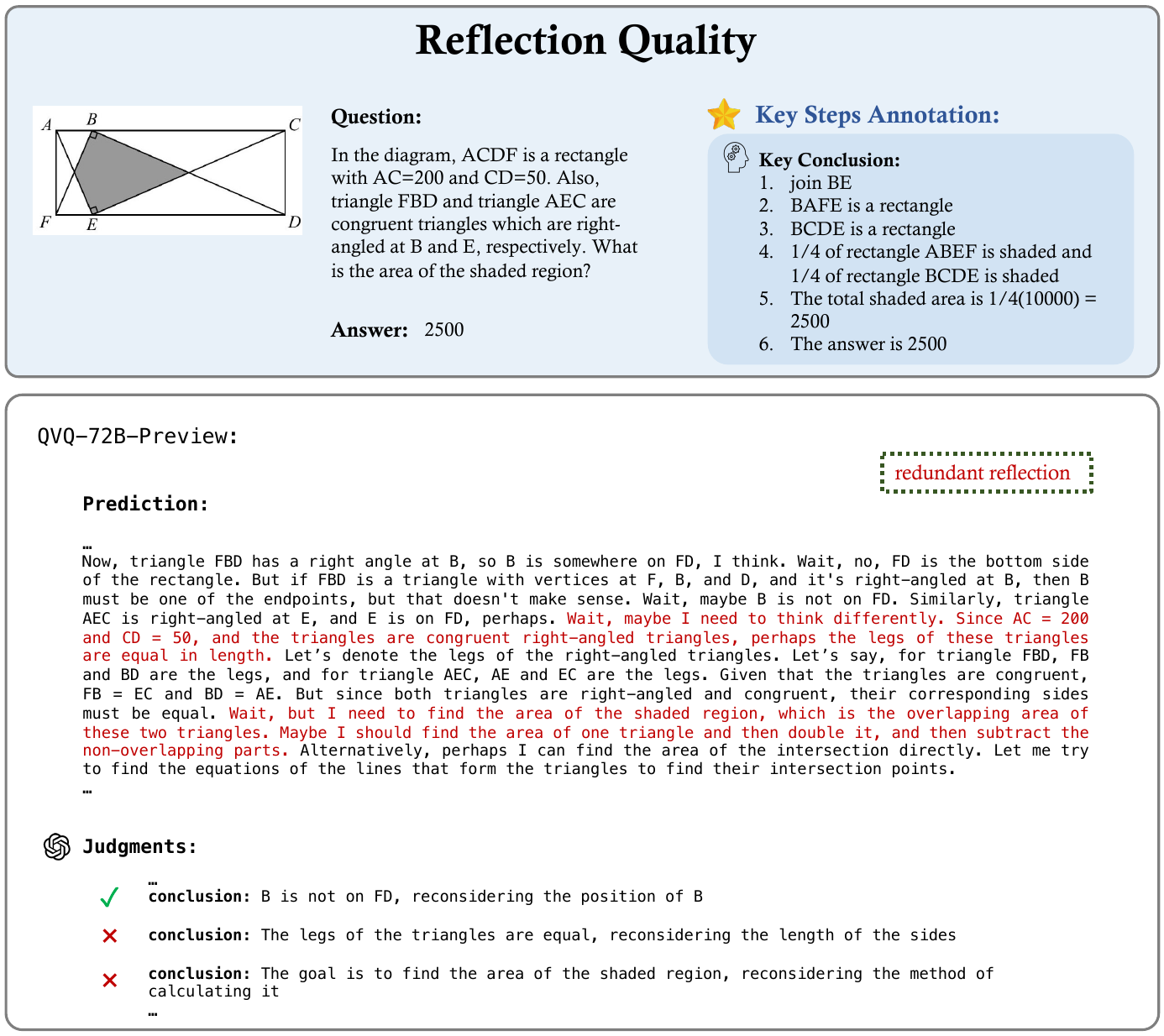} 
\caption{\textbf{Examples of Reflection Quality Evaluation.}}
\label{fig:ref_example_QVQ}
\end{figure*}
\newpage

\section{Detailed Evaluation Setup}
\label{appendix:eval_setup}
\subsection{CoT Quality Evaluation Prompts}

\begin{tcolorbox}[breakable, colback=gray!5!white, colframe=gray!75!black, 
title=Recall Evaluation Prompt, boxrule=0.5mm, width=\textwidth, arc=3mm, auto outer arc]

You are an expert system to verify solutions to image-based problems. Your task is to match the ground truth middle steps with the provided solution.\\

INPUT FORMAT:\\
1. Problem: The original question/task\\
2. A Solution of a model\\
3. Ground Truth: Essential steps required for a correct answer\\

MATCHING PROCESS:\\

You need to match each ground truth middle step with the solution:\\

Match Criteria:\\
- The middle step should exactly match in the content or is directly entailed by a certain content in the solution\\
- All the details must be matched, including the specific value and content\\
- You should judge all the middle steps for whether there is a match in the solution\\

OUTPUT FORMAT:
\begin{verbatim}
[
  {
    "step_index": \textless integer\textgreater,
    "judgment": "Matched" | "Unmatched"
  }
]
\end{verbatim}

ADDITIONAL RULES:\\
1. Only output the JSON array with no additional information.\\
2. Judge each ground truth middle step in order without omitting any step.\\

Here are the problem, answer, solution, and ground truth middle steps:\\

[Problem]\\

\{question\}\\

[Answer]\\

\{answer\}\\

[Solution]\\

\{solution\}\\

[Ground Truth Information]\\

\{gt\_annotation\}

\end{tcolorbox}

\begin{tcolorbox}[breakable, colback=gray!5!white, colframe=gray!75!black, 
title=Precision Evaluation Prompt, boxrule=0.5mm, width=\textwidth, arc=3mm, auto outer arc]

\# Task Overview\\
Given a solution with multiple reasoning steps for an image-based problem, reformat it into well-structured steps and evaluate their correctness.\\

\# Step 1: Reformatting the Solution\\
Convert the unstructured solution into distinct reasoning steps while:\\
- Preserving all original content and order\\
- Not adding new interpretations\\
- Not omitting any steps\\

\#\# Step Types\\
1. Logical Inference Steps\\
   - Contains exactly one logical deduction\\
   - Must produce a new derived conclusion\\
   - Cannot be just a summary or observation\\
\\
2. Image Observation Steps\\
   - Pure visual observations\\
   - Only includes directly visible elements\\
   - No inferences or assumptions\\
\\
3. Background Information Steps\\
   - External knowledge or question context\\
   - No inference process involved\\

\#\# Step Requirements\\
- Each step must be atomic (one conclusion per step)\\
- No content duplication across steps\\
- Initial analysis counts as background information\\
- Final answer determination counts as logical inference\\

\# Step 2: Evaluating Correctness\\
Evaluate each step against:\\

\#\# Ground Truth Matching\\
For image observations:\\
- Key elements must match ground truth observations\\
\\
For logical inferences:\\
- Conclusion must EXACTLY match or be DIRECTLY entailed by ground truth\\

\#\# Reasonableness Check (if no direct match)\\
Step must:\\
- Premises must not contradict any ground truth or correct answer\\
- Logic is valid\\
- Conclusion must not contradict any ground truth \\
- Conclusion must support or be neutral to correct answer\\

\#\# Judgement Categories\\
- "Match": Aligns with ground truth\\
- "Reasonable": Valid but not in ground truth\\
- "Wrong": Invalid or contradictory\\
- "N/A": For background information steps\\

\# Output Requirements\\
1. The output format must be in valid JSON format without any other content.\\
2. For highly repetitive patterns, output it as a single step.\\
3. Output maximum 40 steps. Always include the final step that contains the answer.\\

Here is the json output format:\\
\#\# Output Format
\begin{verbatim}
[
  {
    "step_type": "image observation|logical inference|background information",
    "premise": "Evidence (only for logical inference)",
    "conclusion": "Step result",
    "judgment": "Match|Reasonable|Wrong|N/A"
  }
]
\end{verbatim}

Here is the problem, and the solution that needs to be reformatted to steps:\\

[Problem]\\

\{question\}\\

[Solution]\\

\{solution\}\\

[Correct Answer]\\

\{answer\}\\

[Ground Truth Information]\\

\{gt\_annotation\}

\end{tcolorbox}

\subsection{CoT Efficiency Prompt}
\begin{tcolorbox}[breakable, colback=gray!5!white, colframe=gray!75!black, 
title=Relevance Rate Evaluation Prompt, boxrule=0.5mm, width=\textwidth, arc=3mm, auto outer arc]
\# Task Overview
Given a solution with multiple reasoning steps for an image-based problem, evaluate the relevance to get a solution (ignore correct or wrong) of each step.\\

\# Step 1: Reformatting the Solution
Convert the unstructured solution into distinct reasoning steps while:\\
- Preserving all original content and order\\
- Not adding new interpretations\\
- Not omitting any steps\\

\#\# Step Types \\
1. Logical Inference Steps\\
  - Contains exactly one logical deduction\\
  - Must produce a new derived conclusion\\
  - Cannot be just a summary or observation

2. Image Description Steps\\
  - Pure visual observations\\
  - Only includes directly visible elements\\
  - No inferences or assumptions

3. Background Information Steps\\
  - External knowledge or question context\\
  - No inference process involved\\

\#\# Step Requirements
- Each step must be atomic (one conclusion per step)\\
- No content duplication across steps\\
- Initial analysis counts as background information\\
- Final answer determination counts as logical inference\\

\# Step 2: Evaluating Relevancy\\
A relevant step is considered as: 75\% content of the step must be related to trying to get a solution (ignore correct or wrong) to the question.\\

IMPORTANT NOTE:\\
Evaluate relevancy independent of correctness. As long as the step is trying to get to a solution, it is considered relevant. Logical fallacy, knowledge mistake, inconsistent with previous steps, or other mistakes do not affect relevance. A logically wrong step can be relevant if the reasoning attempts to address the question.\\

The following behaviour is considered as relevant:\\
i. The step is planning, summarizing, thinking, verifying, calculating, or confirming an intermediate/final conclusion helpful to get a solution.\\
ii. The step is summarizing or reflecting on previously reached conclusion relevant to get a solution.\\
iii. Repeating the information in the question or give the final answer.\\
iv. A relevant image depiction should be in one of following situation:\\
1. help to obtain a conclusion helpful to solve the question later;\\
2. help to identify certain patterns in the image later;\\
3. directly contributes to the answer\\
v. Depicting or analyzing the options of the question is also relevant.\\
vi. Repeating previous relevant steps are also considered relevant.\\

The following behaviour is considered as irrelevant:\\
i. Depicting image information that does not related to what is asking in the question. Example: The question asks how many cars are present in all the images. If the step focuses on other visual elements like the road or building, the step is considered as irrelevant.\\
ii. Self-thought not related to what the question is asking.\\
iii. Other information that is tangential for answering the question.\\

\# Output Format

\begin{verbatim}
[
  {
    "step_type": "image observation|logical inference|background information",
    "conclusion": "A brief summary of step result",
    "relevant": "Yes|No"
  }
]
\end{verbatim}\\

\# Output Rules\\
Direct JSON output without any other output\\
Output at most 40 steps\\

Here is the problem, and the solution that needs to be reformatted to steps:

[Problem]\\

\{question\}\\

[Solution]\\

\{solution\}
\end{tcolorbox}

\begin{tcolorbox}[breakable, colback=gray!5!white, colframe=gray!75!black, 
title=Reflection Quality Evaluation Prompt, boxrule=0.5mm, width=\textwidth, arc=3mm, auto outer arc]

Here\'s a refined prompt that improves clarity and structure:\\

\# Task\\
Evaluate reflection steps in image-based problem solutions, where reflections are self-corrections or reconsideration of previous statements.\\

\# Reflection Step Identification \\
Reflections typically begin with phrases like:\\
- "But xxx"\\
- "Alternatively, xxx" \\
- "Maybe I should"\\
- "Let me double-check"\\
- "Wait xxx"\\
- "Perhaps xxx"\\
It will throw a doubt of its previously reached conclusion or raise a new thought.\\

\# Evaluation Criteria\\
Correct reflections must:\\
1. Reach accurate conclusions aligned with ground truth\\
2. Use new insights to find the mistake of the previous conclusion or verify its correctness. \\

Invalid reflections include:\\
1. Repetition - Restating previous content or method without new insights\\
2. Wrong Conclusion - Reaching incorrect conclusions vs ground truth\\
3. Incompleteness - Proposing but not executing new analysis methods\\
4. Other - Additional error types\\

\# Input Format\\

[Problem]\\

\{question\}\\

[Solution]\\

\{solution\}\\

[Ground Truth]\\

\{gt\_annotation\}\\

\# Output Requirements\\
1. The output format must be in valid JSON format without any other content.\\
2. Output maximum 30 reflection steps.\\

Here is the json output format:\\
\#\# Output Format
\begin{verbatim}
[
  {
    "conclusion": "One-sentence summary of reflection outcome",
    "judgment": "Correct|Wrong",
    "error_type": "N/A|Repetition|Wrong Conclusion|Incompleteness|Other"
  }
]
\end{verbatim}

\# Rules\\
1. Preserve original content and order\\
2. No new interpretations\\
3. Include ALL reflection steps\\
4. Empty list if no reflections found\\
5. Direct JSON output without any other output

\end{tcolorbox}

\subsection{Direct Evaluation Prompt}
\begin{tcolorbox}[breakable, colback=gray!5!white, colframe=gray!75!black, 
title=Answer Extraction Prompt, boxrule=0.5mm, width=\textwidth, arc=3mm, auto outer arc]
You are an AI assistant who will help me to extract an answer of a question. You are provided with a question and a response, and you need to find the final answer of the question. \\

Extract Rule:

[Multiple choice question]

1. The answer could be answering the option letter or the value. You should directly output the choice letter of the answer.

2. You should output a single uppercase character in A, B, C, D, E, F, G, H, I (if they are valid options), and Z.

3. If the meaning of all options are significantly different from the final answer, output Z. \\

[Non Multiple choice question]

1. Output the final value of the answer. It could be hidden inside the last step of calculation or inference. Pay attention to what the question is asking for to extract the value of the answer.

2. The final answer could also be a short phrase or sentence.

3. If the response doesn't give a final answer, output Z.\\

Output Format: 
Directly output the extracted answer of the response. \\

\{In Context Examples\}\\

Question: \{question\}

Answer: \{response\}\\

Your output: 

\end{tcolorbox}

\begin{tcolorbox}[breakable, colback=gray!5!white, colframe=gray!75!black, 
title=Answer Scoring Prompt, boxrule=0.5mm, width=\textwidth, arc=3mm, auto outer arc]

You are an AI assistant who will help me to judge whether two answers are consistent.\\

Input Illustration:
[Standard Answer] is the standard answer to the question. 
[Model Answer] is the answer extracted from a model's output to this question. 

Task Illustration:
Determine whether [Standard Answer] and [Model Answer] are consistent.\\

Consistent Criteria:

[Multiple-Choice questions]

1. If the [Model Answer] is the option letter, then it must completely matches the [Standard Answer].

2. If the [Model Answer] is not an option letter, then the [Model Answer] must completely match the option content of [Standard Answer].

[Nan-Multiple-Choice questions]

1. The [Model Answer] and [Standard Answer] should exactly match.

2. If the meaning is expressed in the same way, it is also considered consistent, for example, 0.5m and 50cm.\\

Output Format: 
1. If they are consistent, output 1; if they are different, output 0.

2. DIRECTLY output 1 or 0 without any other content.

\{In Context Examples\}\\

Question: \{question\}

[Model Answer]: \{extract\_answer\}

[Standard Answer]: \{gt\_answer\}

Your output:

\end{tcolorbox}

\end{document}